\pdfoutput=1

\documentclass[11pt]{article}

\usepackage[final]{acl}

\usepackage{times}
\usepackage{latexsym}

\usepackage[T1]{fontenc}

\usepackage[utf8]{inputenc}

\usepackage{microtype}

\usepackage{inconsolata}

\usepackage{graphicx}
\usepackage{amsmath}  
\usepackage{amssymb}  
\usepackage{amsfonts} 
\usepackage{enumitem}
\usepackage{booktabs} 
\usepackage{geometry} 
\usepackage{multirow} 
\usepackage{titletoc}
\usepackage{fontawesome}

\usepackage[most]{tcolorbox}

\tcbset{
  finding style/.style={
    enhanced,
    colback=gray!10,           
    colframe=gray!60,         
    boxrule=0.8pt,             
    arc=2mm,                   
    left=1mm, right=1mm,       
    top=1mm, bottom=1mm,
    before skip=10pt, after skip=10pt,
    fontupper=\normalfont,     
    fonttitle=\normalfont\bfseries, 
  }
}

\newtcolorbox{findingbox}[1][]{finding style,#1}

\title{Can LLMs Estimate Student Struggles? Human-AI Difficulty Alignment with Proficiency Simulation for Item Difficulty Prediction}

\author{ 
  \textbf{Ming Li}\textsuperscript{*1}, 
  \textbf{Han Chen}\thanks{Equal Contribution.}, 
  \textbf{Yunze Xiao}\textsuperscript{2}, 
  \textbf{Jian Chen}\textsuperscript{3}, 
  \textbf{Hong Jiao}\textsuperscript{1}, 
  \textbf{Tianyi Zhou}\textsuperscript{4}\\
  \textsuperscript{1}University of Maryland~~ \textsuperscript{2}Carnegie Mellon University~~ \textsuperscript{3}University at Buffalo~~ \textsuperscript{4}MBZUAI\\
  \texttt{minglii@umd.edu, tianyi.david.zhou@gmail.com} \\
    \faGithub~Project: \url{https://github.com/MingLiiii/Difficulty_Alignment}
}

\begin{document}
\maketitle
\begin{abstract}
    Accurate estimation of item (question or task) difficulty is critical for educational assessment but suffers from the cold start problem. While Large Language Models demonstrate superhuman problem-solving capabilities, it remains an open question whether they can perceive the cognitive struggles of human learners.
    In this work, we present a large-scale empirical analysis of \textbf{Human-AI Difficulty Alignment} for over 20 models across diverse domains such as medical knowledge and mathematical reasoning.
    Our findings reveal a systematic misalignment where scaling up model size is not reliably helpful; instead of aligning with humans, models converge toward a shared machine consensus.
    We observe that high performance often impedes accurate difficulty estimation, as models struggle to simulate the capability limitations of students even when being explicitly prompted to adopt specific proficiency levels.
    Furthermore, we identify a critical lack of introspection, as models fail to predict their own limitations.
    These results suggest that general problem-solving capability does not imply an understanding of human cognitive struggles, highlighting the challenge of using current models for automated difficulty prediction.
    \looseness-1
\end{abstract}

\section{Introduction}

Accurate estimation of item difficulty is the cornerstone of educational assessment \citep{hambleton1991fundamentals, hsu2018automated, alkhuzaey2021systematic, peters2025text}. It underpins critical applications such as curriculum design, automated test generation, and automated item generation with controlled difficulty levels \citep{demars2010item, lord2012applications}. Traditionally, obtaining accurate difficulty parameters (e.g., within Item Response Theory (IRT) models \citep{baker2001basics, lalor2024item}) relies on extensive field testing, a process that requires administering questions to large cohorts of real test-takers to observe response patterns. This reliance creates a significant cold start problem: newly generated questions lack the historical response data necessary to statistically estimate their parameters, effectively rendering them unusable in adaptive systems until they undergo expensive and time-consuming pre-testing cycles.
\looseness-1

Prior approaches to Item Difficulty Prediction (IDP) generally treated the task as a supervised learning problem, relying on linguistic features or deep learning models trained on known item parameters estimated based on item response data \citep{hsu2018automated, benedetto2023quantitative, li2025item}. While effective within specific domains, these methods depend heavily on the availability of historical performance data for training, limiting their utility in cold-start scenarios (i.e., no historical tested data is available for training). The emergence of LLMs \citep{openai_chatgpt_whisper_api_2024, hurst2024gpt, touvron2023llama,qwen2.5,qwen3technicalreport} offers a potential paradigm shift. With their vast pre-training and exceptional problem-solving capabilities, LLMs seemingly possess the knowledge required to analyze complex content. \textbf{\textit{However, it remains an open question whether these general-purpose models can align with human perception of difficulty without task-specific fine-tuning}}. There is a fundamental distinction between solving a problem and evaluating its difficulty: a model that effortlessly surpasses human baselines in performance may fail to recognize the cognitive hurdles faced by an average learner \citep{sweller1988cognitive, sweller2011cognitive, noroozi2022scrutiny, li-etal-2025-understanding}. This study investigates this \textit{\textbf{Human-AI Difficulty Alignment}}, exploring whether off-the-shelf LLMs can bridge the gap between their own capabilities and the student struggles, whose difficulty values are obtained from real student field testing.

To investigate this, we propose a comprehensive empirical study that evaluates this Difficulty Alignment through two distinct lenses: the model as an \textbf{external observer} (predicting others' difficulty) and an \textbf{internal actor} (experiencing difficulty itself). Our study operates at scale, benchmarking over $20$ LLMs, spanning both open-weights and closed-source families, including reasoning-specialized models, across four diverse educational domains: language proficiency (Cambridge) \citep{cambridge_reading_comprehension}, reasoning and logic (SAT Reading/Writing, SAT Math), and professional medical knowledge (USMLE) \citep{yaneva2024findings}.

We structure our investigation around three primary dimensions to disentangle the relationship between intrinsic capability and extrinsic perception. \textbf{\textit{First}}, we go beyond simple ground-truth correlation to analyze inter-model consensus, examining whether models form a cohesive machine perception that systematically diverges from students. \textbf{\textit{Second}}, we quantify the capability-perception gap using \textbf{Item Response Theory (IRT)}. By treating the model pool as a cohort of synthetic students, we derive empirical machine difficulty based on the actual correctness of LLMs, allowing us to test for the Curse of Knowledge, where items challenging for humans are trivial for machines. \textbf{\textit{Finally}}, we evaluate \textbf{Metacognitive Alignment} and \textbf{Proficiency Simulation} \citep{tseng2024two, zhang2024personalization, hayakawa2024can}, rigorously testing whether models possess the introspection to predict their own limitations or the flexibility to authentically simulate the cognitive struggles of lower-proficiency students.

\vspace{-1mm}
\paragraph{Key Findings.}
\begin{enumerate}[leftmargin=4mm]
    \vspace{-1mm}
    \item \textbf{Systematic Misalignment:} Contrary to standard capability metrics, scaling does not reliably translate into alignment. Increasing model scale does not improve difficulty predictions; instead, models form a cohesive Machine Consensus, aligning significantly stronger with each other than with human reality.

    \vspace{-1mm}
    \item \textbf{Limits of Simulation:} Neither extrinsic ensembling nor proficiency simulation serves as a reliable fix for the misalignment. Ensemble performance is strictly bounded by weaker models, while proficiency simulation proves highly inconsistent as models struggle to authentically mimic different proficiency levels.

    \vspace{-1mm}
    \item \textbf{The Curse of Knowledge:} Our IRT-based analysis reveals a fundamental mechanistic divergence: the difficulty derived from models' actual correctness correlates even worse with humans than their explicit perceptions. Items that are difficult for humans are frequently trivial for models, and this capability exhibits significant inertia even under weak student prompts.

    \vspace{-1mm}
    \item \textbf{Metacognitive Blindness:} We identify a critical lack of introspection. With AUROC scores hovering near random guessing, models fail to predict their own limitations, indicating that explicit difficulty estimates are effectively decoupled from the model's actual correctness, lacking the internal signal to ground their predictions.
\end{enumerate}

\section{Formulation and Evaluation Metrics}
\label{sec:formulation}

\subsection{Task Formulation}
We formalize the IDP task as a function approximation problem over a dataset $\mathcal{D} = \{(x_i, a^*_i, y_i)\}_{i=1}^N$. Each entry comprises the item context $x_i$ (including the question stem, optional passages, and candidate options), the ground truth answer $a^*_i$, and the difficulty label $y_i \in \mathcal{Y}$ obtained from real-world student field testing. The label space $\mathcal{Y}$ adapts to the domain, ranging from continuous values (e.g., $y_i \in [0, 1]$) to discrete categories (e.g., $\mathcal{Y} = \{\text{Easy, Medium, Hard}\}$).

Let $\mathcal{M}$ denote the set of LLMs under evaluation. We investigate the alignment between human difficulty $y_i$ and AI cognition through two distinct modalities: \textit{Difficulty Perception} (the model's estimation as an observer) and \textit{Problem-Solving Capability} (the model's performance as an actor).

\paragraph{The Observer View: Difficulty Perception.}
In this mode, the model simulates an educator or test developer tasked with estimating item difficulty. To isolate perception from solving capability, the model is provided with the full item context $x_i$, the correct solution $a^*_i$, and an optional proficiency prompt $p$. The predicted difficulty $\hat{y}_{i,m}$ for model $m \in \mathcal{M}$ is formulated as:
\begin{equation}
    \hat{y}_{i,m} = \phi\left( \text{Gen}_m(x_i, a^*_i, p) \right)
\end{equation}
where $\text{Gen}_m(\cdot)$ denotes the natural language generation process, and $\phi(\cdot)$ is a parsing function that maps the generated response to a normalized numerical difficulty score.

\paragraph{The Actor View: Intrinsic Capability.}
To assess the model's actual performance, we evaluate it in a standard zero-shot test-taking setting where the correct answer is hidden. The model generates a solution:
\begin{equation}
    \hat{a}_{i,m} = \psi( \text{Gen}_m(x_i, p) )
\end{equation}
where $\psi(\cdot)$ extracts the final answer. The binary correctness $v_{i,m} \in \{0, 1\}$ is subsequently determined by $v_{i,m} = \mathbb{I}(\hat{a}_{i,m} = a^*_i)$.

\subsection{Evaluation Framework}
\label{sec:metrics}

To analyze the divergence between human and AI cognition, we evaluate alignment across two distinct dimensions: \textit{Perceived Difficulty} (what models predict) and \textit{Empirical Difficulty} (what models experience). We adopt \textbf{Spearman's Rank Correlation ($\rho$)} as the unified metric for both dimensions, as it robustly measures monotonicity, the ability to correctly distinguish that item A is harder than item B, while remaining invariant to systematic scaling shifts. Moreover, Spearman correlation allows unified comparison across heterogeneous label granularities, discrete or continuous.
\looseness-1

\paragraph{Perception Alignment ($\rho_{pred}$).}
This metric evaluates the model's accuracy as an \textit{observer}. We calculate the Spearman correlation between the model's predicted difficulty scores $\hat{y}_{i,m}$ and the human ground truth $y_i$.
\begin{equation}
    \rho_{pred,m} = \text{Spearman}(\{\hat{y}_{i,m}\}_{i=1}^N, \{y_i\}_{i=1}^N)
\end{equation}
A higher $\rho_{pred,m}$ indicates that the model's explicit perception of difficulty aligns with the human hierarchy. For simplicity, we omit the subscript $m$ when not referring to a specific model.

\paragraph{Capability Alignment ($\rho_{irt}$).}
This metric evaluates the model's alignment as an \textit{actor}. To quantify this, we first derive the \textit{Empirical Machine Difficulty} $\beta_i$ using Item Response Theory (IRT). We treat the set of models $\mathcal{M}$ as a population of synthetic examinees and construct a binary correctness matrix. We fit a Rasch Model (1-Parameter Logistic Model) where the probability of model $m$ answering item $i$ correctly is:
\begin{equation}
    P(v_{i,m}=1 \mid \theta_{m}, \beta_{i}) = \frac{1}{1 + \exp(-(\theta_{m} - \beta_{i}))}
\end{equation}
Here, $\beta_{i}$ represents the intrinsic machine difficulty of item $i$, estimated via Marginal Maximum Likelihood Estimation.
Crucially, we then calculate the correlation between this empirical difficulty and human difficulty:
\begin{equation}
    \rho_{irt} = \text{Spearman}(\{\beta_i\}_{i=1}^N, \{y_i\}_{i=1}^N)
\end{equation}
With $\rho_{pred}$ and $\rho_{irt}$, we can obtain a systemic view of the Human-AI Difficulty Alignment.

\subsection{Proficiency Simulation}
To systematically investigate the model's ability to simulate different cognitive states and align with student populations, we define a set of four distinct proficiency configurations $\mathcal{P} = \{p_0, p_{low}, p_{mid}, p_{high}\}$. These configurations serve as system-level instructions that condition the model's generation process.

\begin{itemize}[leftmargin=4mm]
    \vspace{-1mm}
    \item \textbf{Baseline: No Proficiency ($p_0$)}. This represents the control setting (or vanilla mode). We provide the model with standard instructions to predict difficulty or solve the problem without adopting a specific student proficiency. This setting evaluates the model's \textit{intrinsic} alignment, i.e., its default perception of difficulty.

    \vspace{-1mm}
    \item \textbf{Low-Proficiency Student ($p_{low}$)}. We prompt the model to simulate a student with limited subject mastery. This proficiency aims to test whether the model can suppress its own knowledge to accurately estimate high-difficulty items for struggling learners.

    \vspace{-1mm}
    \item \textbf{Average-Proficiency Student ($p_{mid}$)}. This proficiency represents the median student. This serves as a proxy for the general population average.

    \vspace{-1mm}
    \item \textbf{High-Proficiency Student ($p_{high}$)}. We simulate a top-tier student who has high proficiency in the subject. This setting investigates whether the model aligns closely with the most capable subset of the human population.
\end{itemize}

For different domains, the proficiency prompts are slightly different to align with the domain-specific scenarios, as shown in the Appendix \ref{sec:experimental_prompts}.
Moreover, we do not provide much description on what a low/average/high-proficiency student would be like, since this is also a part of the investigation on how the model understands them.

\begin{table}[t]
    \centering
    \setlength{\tabcolsep}{2.5pt}
    \resizebox{\linewidth}{!}{
        \begin{tabular}{l|cccc|c}
        \toprule
        \textbf{Model} & \textbf{USMLE} & \textbf{Cambridge} & \textbf{SAT-R} & \textbf{SAT-M} & \textbf{Average} \\
        \midrule
        GPT-3.5-Turbo   & 0.09 & 0.20 & 0.26 & 0.40 & 0.24 \\
        GPT-4o          & 0.19 & \underline{0.48} & 0.38 & \underline{0.55} & 0.40 \\
        GPT-4o-mini     & 0.05 & 0.42 & 0.26 & 0.49 & 0.31 \\
        GPT-4.1         & \underline{0.30} & \textbf{0.49} & \underline{0.49} & 0.48 & \textbf{0.44} \\
        GPT-4.1-mini    & 0.20 & 0.46 & 0.48 & 0.45 & 0.40 \\
        GPT-o4-mini     & 0.28 & 0.36 & 0.46 & 0.29 & 0.35 \\
        GPT-5           & \textbf{0.36} & 0.36 & 0.38 & 0.25 & 0.34 \\
        \midrule
        Llama2-7B       & 0.05 & 0.00 & 0.08 & 0.03 & 0.04 \\
        Llama2-13B      & 0.00 & 0.10 & 0.14 & 0.16 & 0.10 \\
        Phi3            & 0.01 & -0.01 & 0.09 & 0.30 & 0.10 \\
        Phi3.5          & 0.06 & 0.06 & 0.12 & 0.32 & 0.14 \\
        Llama3.1-8B     & 0.04 & 0.04 & 0.23 & 0.43 & 0.19 \\
        Qwen2.5-7B      & 0.10 & 0.16 & 0.22 & 0.49 & 0.24 \\
        Qwen2.5-32B     & 0.09 & 0.44 & 0.30 & \textbf{0.56} & 0.35 \\
        Phi4            & 0.11 & 0.40 & 0.38 & 0.50 & 0.35 \\
        Qwen3-8B        & 0.03 & 0.35 & 0.20 & 0.46 & 0.26 \\
        Qwen3-32B       & 0.11 & 0.38 & 0.23 & 0.52 & 0.31 \\
        \midrule
        DeepSeek-R1     & 0.28 & 0.40 & \textbf{0.50} & 0.44 & \underline{0.40} \\
        QWQ-32B         & 0.20 & 0.41 & 0.43 & 0.52 & 0.39 \\
        R1-Qwen32B      & -0.01 & 0.40 & 0.25 & 0.42 & 0.27 \\
        Qwen3-32B (R)   & 0.21 & 0.42 & 0.33 & 0.48 & 0.36 \\
        \midrule
        Average         & 0.13 & 0.30 & 0.29 & 0.41 & 0.28 \\
        \bottomrule
        \end{tabular}

    }
    \vspace{-1.2mm}
    \caption{
        The Spearman's rank correlation results. \textbf{Overall alignment remains weak, indicating systematic misalignment.}
    }
    \vspace{-2.2mm}
    \label{tab:table_diff_by_task}
\end{table}

\subsection{Experimental Setup}

\paragraph{Datasets.}
The key challenge for Human-AI Difficulty Alignment is to find the dataset that has the ground truth difficulty values that are obtained from real student field testing, since most of the existing research on IDP is conducted on private datasets or datasets from multiple resources, resulting really large discrepancy between the model's capability and the real students' performance.
To ensure the robustness of our findings across different educational domains, we select four datasets covering reading comprehension, verbal reasoning, math reasoning, and specialized professional knowledge.

\vspace{-1mm}
\paragraph{USMLE (Medical Knowledge):} Sourced from the United States Medical Licensing Examination \cite{yaneva2024findings}, this dataset represents a high-stakes, knowledge-intensive domain. It contains 667 items developed by the NBME and FSMB, ensuring high reliability with field-test data from over 300 medical students per item. We utilize the provided continuous difficulty values (transformed p-values), which range from $[0, 1.3]$.

\vspace{-1mm}
\paragraph{Cambridge (Linguistic Proficiency):} Sourced from the Cambridge Multiple-Choice Questions Reading Dataset \cite{cambridge_reading_comprehension}, this dataset evaluates English reading comprehension through 120 text passages and 793 distinct questions. A key characteristic is the long context window required for inference. We utilize the rescaled IRT $b$-parameters provided in the dataset as ground truth, which represent continuous difficulty values in the range of $[0, 100]$.

\vspace{-1mm}
\paragraph{SAT Reading \& Writing (Verbal Reasoning):} Comprises reading comprehension and writing mechanics questions from standardized US college admission tests. This dataset challenges the model's ability to process rhetorical structure and standard English conventions. The dataset contains $1338$ questions after removing the figure-dependent questions. Unlike the datasets above, the ground truth difficulty is provided as discrete categories: $\{\textit{Easy}, \textit{Medium}, \textit{Hard}\}$.
\vspace{-1mm}

\paragraph{SAT Math (Mathematical Logic):} Includes algebra, geometry, and data analysis problems from the SAT. This dataset tests the model's ability to gauge difficulty in multi-step logical reasoning and mathematical computation tasks. Similar to the verbal component, it contains $1385$ questions, and the difficulty values are provided as discrete categories: $\{\textit{Easy}, \textit{Medium}, \textit{Hard}\}$.

\begin{figure}[t]
    \centering
    \includegraphics[width=0.46\textwidth]{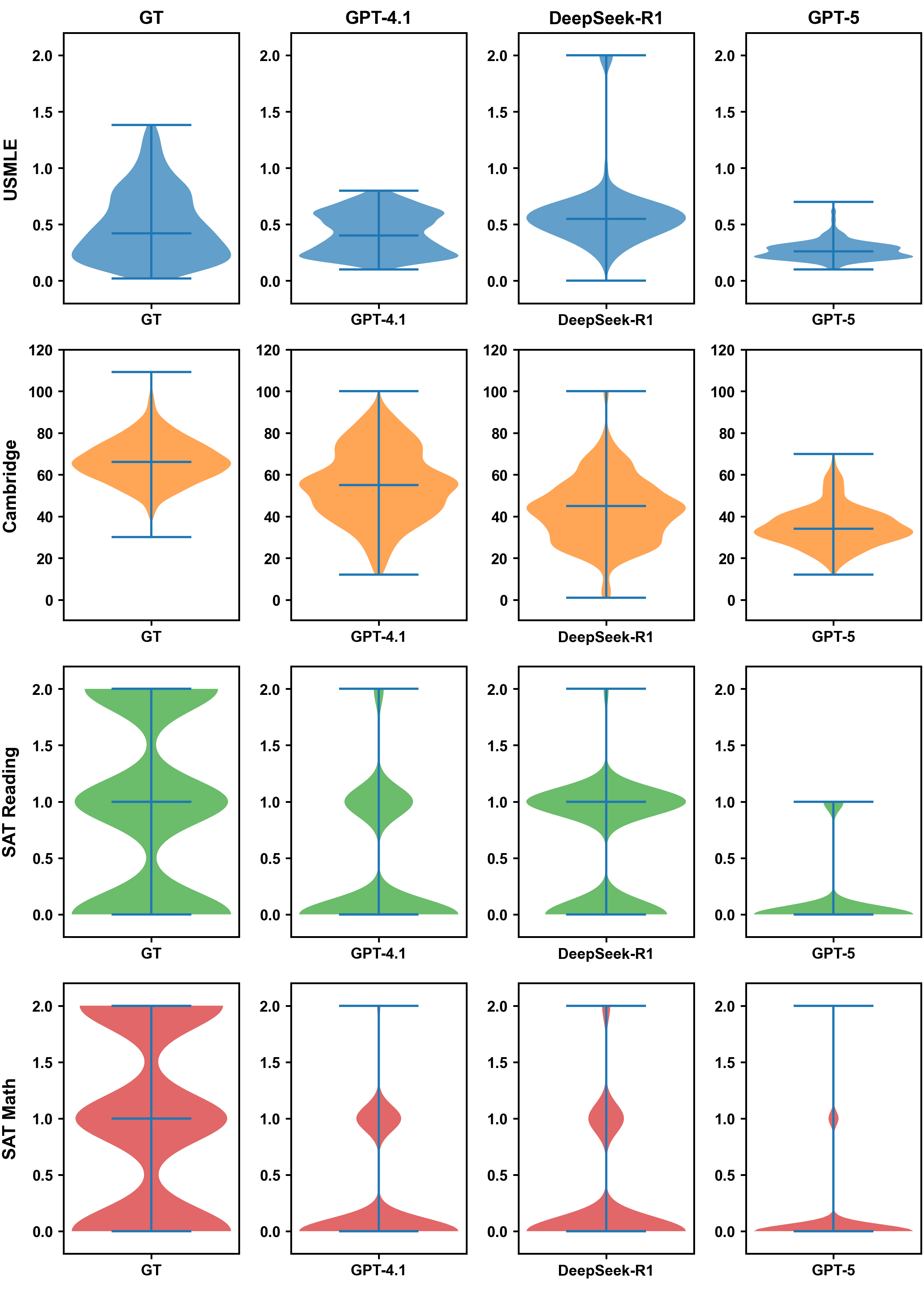}
    \vspace{-3.2mm}
    \caption{
    The violin plot of the difficulty prediction distributions of several representative models. \textbf{Current advanced models exhibit severe \textit{distribution shift}. }
    }
    \vspace{-3.2mm}
    \label{fig:figure_diff_by_task_violin}
    \end{figure}

\paragraph{Models.}
We evaluate a comprehensive suite of over $20$ LLMs to disentangle the effects of models.
We establish a high-performance baseline using proprietary models, including the GPT series (GPT-3.5-Turbo~\citep{openai_chatgpt_whisper_api_2024}, GPT-4o~\citep{hurst2024gpt}, GPT-4o-mini~\citep{openai_gpt4o_mini_2024}, GPT-4.1~\cite{openai_gpt4_1_2025}, GPT-4.1-mini~\cite{openai_gpt4_1_2025}, GPT-o4-mini~\cite{openai_o3_o4mini_2024}, GPT-5~\cite{openai_gpt5_system_card_2025}), which represent the current state-of-the-art in general-purpose tasks.
The general instruction-following open-weights models including Llama2-7B~\citep{touvron2023llama}, Llama2-13B~\citep{touvron2023llama}, Llama3.1-8B~\citep{dubey2024llama}, Qwen2.5-7B~\citep{qwen2.5}, Qwen2.5-32B~\citep{qwen2.5}, Phi3~\citep{abdin2024phi3technicalreporthighly}, Phi3.5~\citep{abdin2024phi3technicalreporthighly}, Phi4~\citep{abdin2024phi}, Qwen3-8B, and Qwen3-32B~\citep{qwen3technicalreport}.
To specifically investigate the impact of reasoning capabilities, we incorporate reasoning-focused open-weights models, including DeepSeek-R1(0528)~\citep{deepseekai2025deepseekr1incentivizingreasoningcapability}, QWQ-32B~\citep{qwq32b}, R1-Distill-Qwen2.5-32B~\citep{deepseekai2025deepseekr1incentivizingreasoningcapability}, and Qwen3-32B (reasoning mode).

\section{The Landscape of Explicit Perception}

We first establish a baseline for zero-shot IDP, examining whether current state-of-the-art LLMs can intrinsically estimate item difficulty without access to student response data. The results, summarized in Table \ref{tab:table_diff_by_task}, reveal three primary findings regarding the current limitations of Human-AI difficulty alignment.

\subsection{Systematic Misalignment}

\begin{figure}[t]
    \centering
    \includegraphics[width=0.46\textwidth]{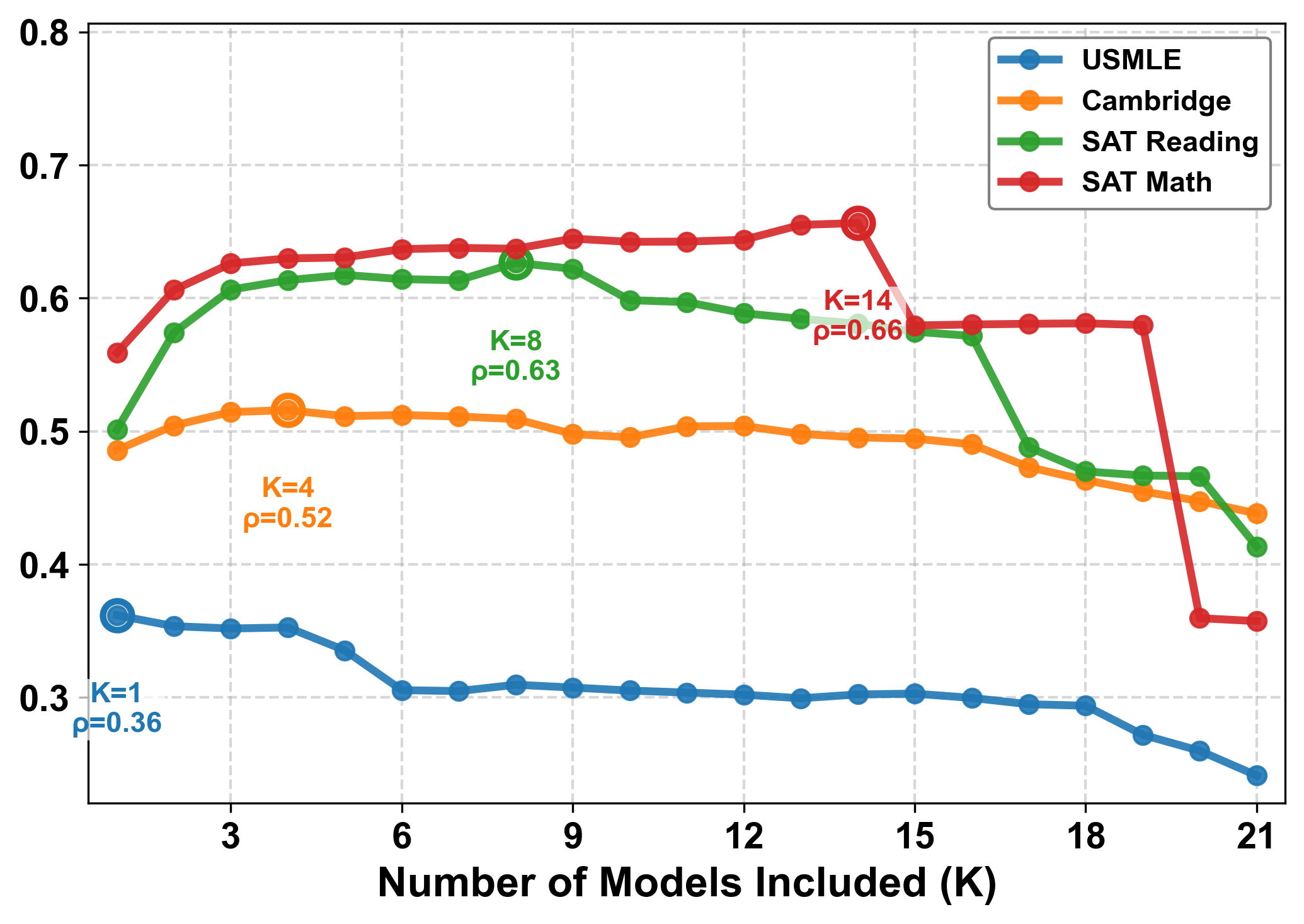}
    \vspace{-3.2mm}
    \caption{
        The Spearman correlation trends when greedily ensembling the predictions of the top-$K$ models. \textbf{The curve indicates the upper bound of the ensemble performance, which is still weak.}
    }
    \vspace{-3.2mm}
    \label{fig:ensemble_topk}
    \end{figure}

\begin{figure}[t]
    \centering
    \includegraphics[width=0.48\textwidth]{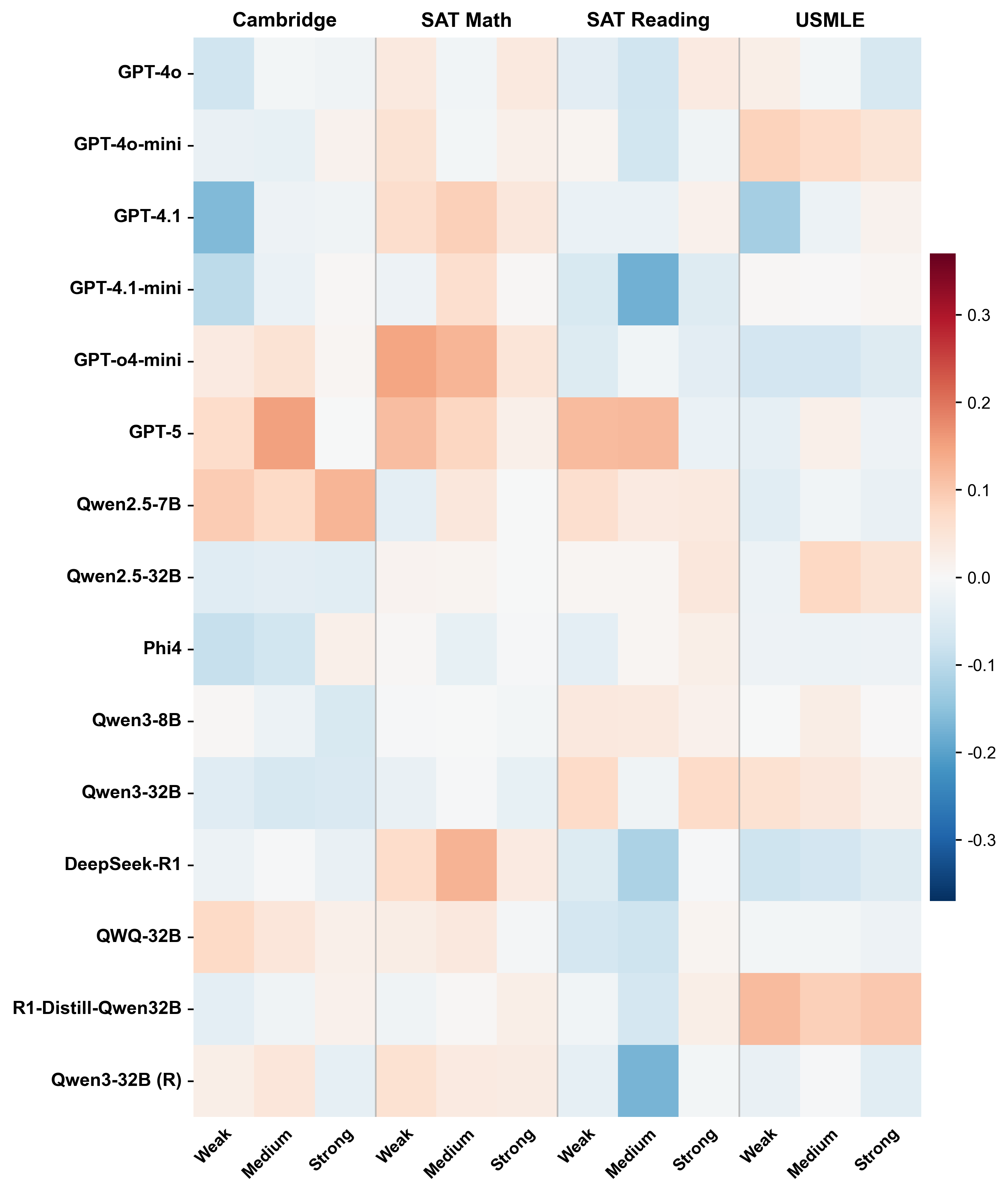}
    \vspace{-5.2mm}
    \caption{
        Heatmap showing the correlation change when applying specific personas compared to the baseline. \textbf{The impact of individual personas is highly inconsistent and noisy.
    } }
    \vspace{-3.2mm}
    \label{fig:hetmap_role_diff}
    \end{figure}

\begin{table}[t]
    \centering
    \setlength{\tabcolsep}{2.5pt}
    \resizebox{\linewidth}{!}{
        \begin{tabular}{l|cccc|c}
        \toprule
        \textbf{Model} & \textbf{USMLE} & \textbf{Cambridge} & \textbf{SAT-R} & \textbf{SAT-M} & \textbf{Average} \\
        \midrule
        GPT-3.5-Turbo   & 0.14 {\scriptsize \textcolor{red}{(0.05)}} & 0.32 {\scriptsize \textcolor{red}{(0.12)}} & 0.31 {\scriptsize \textcolor{red}{(0.06)}} & 0.53 {\scriptsize \textcolor{red}{(0.13)}} & 0.33 {\scriptsize \textcolor{red}{(0.09)}} \\
        GPT-4o          & 0.22 {\scriptsize \textcolor{red}{(0.04)}} & 0.50 {\scriptsize \textcolor{red}{(0.02)}} & 0.49 {\scriptsize \textcolor{red}{(0.11)}} & \textbf{0.65} {\scriptsize \textcolor{red}{(0.09)}} & 0.47 {\scriptsize \textcolor{red}{(0.07)}} \\
        GPT-4o-mini     & 0.12 {\scriptsize \textcolor{red}{(0.07)}} & 0.46 {\scriptsize \textcolor{red}{(0.04)}} & 0.35 {\scriptsize \textcolor{red}{(0.09)}} & 0.57 {\scriptsize \textcolor{red}{(0.08)}} & 0.38 {\scriptsize \textcolor{red}{(0.07)}} \\
        GPT-4.1         & \underline{0.31} {\scriptsize \textcolor{red}{(0.01)}} & \underline{0.51} {\scriptsize \textcolor{red}{(0.02)}} & \textbf{0.60} {\scriptsize \textcolor{red}{(0.11)}} & 0.63 {\scriptsize \textcolor{red}{(0.15)}} & \textbf{0.51} {\scriptsize \textcolor{red}{(0.07)}} \\
        GPT-4.1-mini    & 0.23 {\scriptsize \textcolor{red}{(0.03)}} & 0.49 {\scriptsize \textcolor{red}{(0.03)}} & 0.56 {\scriptsize \textcolor{red}{(0.08)}} & 0.56 {\scriptsize \textcolor{red}{(0.11)}} & 0.46 {\scriptsize \textcolor{red}{(0.06)}} \\
        GPT-o4-mini     & 0.26 {\scriptsize (-0.02)} & 0.46 {\scriptsize \textcolor{red}{(0.10)}} & 0.56 {\scriptsize \textcolor{red}{(0.10)}} & 0.49 {\scriptsize \textcolor{red}{(0.20)}} & 0.44 {\scriptsize \textcolor{red}{(0.10)}} \\
        GPT-5           & \textbf{0.39} {\scriptsize \textcolor{red}{(0.03)}} & \textbf{0.52} {\scriptsize \textcolor{red}{(0.16)}} & 0.57 {\scriptsize \textcolor{red}{(0.19)}} & 0.40 {\scriptsize \textcolor{red}{(0.15)}} & \underline{0.47} {\scriptsize \textcolor{red}{(0.13)}} \\
        \midrule
        Llama2-7B       & 0.09 {\scriptsize \textcolor{red}{(0.02)}} & -0.03 {\scriptsize (-0.03)} & 0.10 {\scriptsize \textcolor{red}{(0.03)}} & 0.14 {\scriptsize \textcolor{red}{(0.11)}} & 0.08 {\scriptsize \textcolor{red}{(0.03)}} \\
        Llama2-13B      & 0.03 {\scriptsize \textcolor{red}{(0.03)}} & 0.02 {\scriptsize (-0.06)} & 0.12 {\scriptsize (-0.02)} & 0.19 {\scriptsize \textcolor{red}{(0.03)}} & 0.09 {\scriptsize (-0.01)} \\
        Phi3            & 0.04 {\scriptsize \textcolor{red}{(0.04)}} & 0.08 {\scriptsize \textcolor{red}{(0.09)}} & 0.07 {\scriptsize \textcolor{red}{(0.01)}} & 0.01 {\scriptsize (-0.04)} & 0.05 {\scriptsize \textcolor{red}{(0.03)}} \\
        Phi3.5          & 0.02 {\scriptsize (-0.03)} & 0.17 {\scriptsize \textcolor{red}{(0.12)}} & 0.18 {\scriptsize \textcolor{red}{(0.03)}} & 0.35 {\scriptsize \textcolor{red}{(0.03)}} & 0.18 {\scriptsize \textcolor{red}{(0.04)}} \\
        Llama3.1-8B     & 0.01 {\scriptsize (-0.03)} & 0.20 {\scriptsize \textcolor{red}{(0.16)}} & 0.33 {\scriptsize \textcolor{red}{(0.10)}} & 0.55 {\scriptsize \textcolor{red}{(0.11)}} & 0.27 {\scriptsize \textcolor{red}{(0.09)}} \\
        Qwen2.5-7B      & 0.13 {\scriptsize \textcolor{red}{(0.03)}} & 0.34 {\scriptsize \textcolor{red}{(0.18)}} & 0.36 {\scriptsize \textcolor{red}{(0.14)}} & 0.63 {\scriptsize \textcolor{red}{(0.13)}} & 0.37 {\scriptsize \textcolor{red}{(0.12)}} \\
        Qwen2.5-32B     & 0.16 {\scriptsize \textcolor{red}{(0.07)}} & 0.47 {\scriptsize \textcolor{red}{(0.04)}} & 0.41 {\scriptsize \textcolor{red}{(0.12)}} & \underline{0.64} {\scriptsize \textcolor{red}{(0.08)}} & 0.42 {\scriptsize \textcolor{red}{(0.08)}} \\
        Phi4            & 0.14 {\scriptsize \textcolor{red}{(0.03)}} & 0.45 {\scriptsize \textcolor{red}{(0.05)}} & 0.47 {\scriptsize \textcolor{red}{(0.09)}} & 0.56 {\scriptsize \textcolor{red}{(0.06)}} & 0.41 {\scriptsize \textcolor{red}{(0.06)}} \\
        Qwen3-8B        & 0.06 {\scriptsize \textcolor{red}{(0.04)}} & 0.41 {\scriptsize \textcolor{red}{(0.07)}} & 0.31 {\scriptsize \textcolor{red}{(0.11)}} & 0.56 {\scriptsize \textcolor{red}{(0.10)}} & 0.34 {\scriptsize \textcolor{red}{(0.08)}} \\
        Qwen3-32B       & 0.19 {\scriptsize \textcolor{red}{(0.08)}} & 0.43 {\scriptsize \textcolor{red}{(0.05)}} & 0.39 {\scriptsize \textcolor{red}{(0.16)}} & 0.61 {\scriptsize \textcolor{red}{(0.09)}} & 0.41 {\scriptsize \textcolor{red}{(0.10)}} \\
        \midrule
        DeepSeek-R1     & 0.26 {\scriptsize (-0.01)} & 0.42 {\scriptsize \textcolor{red}{(0.02)}} & \underline{0.59} {\scriptsize \textcolor{red}{(0.09)}} & 0.62 {\scriptsize \textcolor{red}{(0.18)}} & \underline{0.47} {\scriptsize \textcolor{red}{(0.07)}} \\
        QWQ-32B         & 0.23 {\scriptsize \textcolor{red}{(0.04)}} & 0.46 {\scriptsize \textcolor{red}{(0.11)}} & 0.50 {\scriptsize \textcolor{red}{(0.08)}} & 0.62 {\scriptsize \textcolor{red}{(0.08)}} & 0.45 {\scriptsize \textcolor{red}{(0.08)}} \\
        R1-Qwen32B      & 0.07 {\scriptsize \textcolor{red}{(0.08)}} & 0.46 {\scriptsize \textcolor{red}{(0.06)}} & 0.17 {\scriptsize \textcolor{red}{(0.02)}} & 0.35 {\scriptsize (0.00)} & 0.26 {\scriptsize \textcolor{red}{(0.04)}} \\
        Qwen3-32B (R)   & 0.23 {\scriptsize \textcolor{red}{(0.04)}} & 0.45 {\scriptsize \textcolor{red}{(0.06)}} & 0.37 {\scriptsize \textcolor{red}{(0.10)}} & 0.59 {\scriptsize \textcolor{red}{(0.08)}} & 0.41 {\scriptsize \textcolor{red}{(0.07)}} \\
        \midrule
        Average         & 0.16 {\scriptsize \textcolor{red}{(0.03)}} & 0.36 {\scriptsize \textcolor{red}{(0.07)}} & 0.37 {\scriptsize \textcolor{red}{(0.09)}} & 0.49 {\scriptsize \textcolor{red}{(0.10)}} & 0.35 {\scriptsize \textcolor{red}{(0.07)}} \\
        \bottomrule
        \end{tabular}
    }
    \vspace{-2.2mm}
    \caption{
        Spearman correlation results using the ensemble average of all persona configurations. The values in red parentheses indicate the absolute improvement over the baseline. \textbf{While smaller models show negligible gains, stronger models like GPT-5 exhibit a significant sensitivity to personas.}
    }
    \vspace{-4.2mm}
    \label{tab:table_diff_avg_by_task}
\end{table}

\begin{findingbox}[title={Key Finding 1: Systematic Misalignment.}]
    We observe a \textbf{systematic misalignment} between human perception and LLM estimation of difficulty across all domains. Contrary to standard capability metrics, \textbf{scaling does not reliably translate into alignment}: increased model scale and reasoning power do not linearly translate to better predictions, as even frontier models consistently fail to capture human difficulty rankings. Crucially, this misalignment is not driven by random noise but by a cohesive Machine Consensus, where \textbf{models exhibit relatively stronger alignment with each other than with human reality}.
\end{findingbox}

Despite their profound problem-solving capabilities, LLMs struggle with cold-started IDP. As shown in Table \ref{tab:table_diff_by_task}, Spearman's $\rho$ averages below $0.50$, with significant domain sensitivity: alignment is stronger in logic-driven tasks like \textbf{SAT Math} ($\rho \approx 0.41$) but collapses in knowledge-intensive domains like \textbf{USMLE} ($\rho \approx 0.13$). Counter-intuitively, \textbf{scaling laws do not linearly translate to difficulty estimation}. New-generation models (e.g., GPT-5, $\rho=0.34$) and reasoning specialists fail to outperform generalist baselines like GPT-4.1 ($\rho=0.44$), indicating that reasoning power does not linearly translate to better difficulty alignment with humans.

Figure \ref{fig:figure_diff_by_task_violin} identifies the root cause: severe \textbf{distribution shift and variance collapse}. While ground truth difficulty spans a broad spectrum, model predictions are narrowly clustered and systematically skewed towards lower values. Advanced models effectively \textbf{overestimate student capability}, lacking the granularity to distinguish specific degrees of human struggle.

Finally, we observe a \textbf{Machine Consensus} that diverges from human reality. While alignment with ground truth is low, inter-model correlations are relatively higher (indicated by the detailed heatmaps in Appendix \ref{sec:consensus_of_machines} Figure \ref{fig:figure_appendix_consensus_USMLE}, \ref{fig:figure_appendix_consensus_Cambridge}, \ref{fig:figure_appendix_consensus_SAT_math}, \ref{fig:figure_appendix_consensus_SAT_reading}). This consensus is capability-dependent: while weaker models exhibit stochastic behavior, \textbf{advanced models converge on a shared, non-human machine perception of difficulty}. This confirms that the Human-AI Difficulty Alignment is a stable, systematic misalignment rather than random error.

\subsection{Ensemble and Proficiency Simulation}

\begin{table}[t]
    \centering
    \small
    \setlength{\tabcolsep}{5pt}

    \resizebox{\linewidth}{!}{
    \begin{tabular}{lccccc}
    \toprule
    \multirow{3}{*}{\textbf{Domain}} & \multicolumn{1}{c}{\textbf{IRT}} & \multicolumn{1}{c}{\textbf{Capacity}} & \multicolumn{2}{c}{\textbf{Cognitive Divergence}} \\
    \cmidrule(lr){2-2} \cmidrule(lr){3-3} \cmidrule(lr){4-5}

    & \textbf{Global} & \textbf{Saturated} & \textbf{Savant} & \textbf{Brittle} \\
    & ($\rho$) & (90\% $\mathcal{M}$ \checkmark) & ($\mathcal{H}\!\Uparrow \mathcal{M}\!\Downarrow$) & ($\mathcal{H}\!\Downarrow \mathcal{M}\!\Uparrow$) \\

    \midrule
    \textbf{USMLE} & 0.134 & 75.6\% & 70.4\% & 0.0\% \\
    \textbf{Cambridge}  & 0.309 & 35.6\% &22.1\% & 0.4\% \\
    \textbf{SAT-R} & 0.304 & 45.5\% & 25.5\% & 0.0\% \\
    \textbf{SAT-M} & 0.386 & 54.6\% & 32.2\% & 1.3\% \\
    \bottomrule
    \end{tabular}
    }
    \caption{
    Analysis of implicit difficulty alignment based on Machine IRT.
    $\mathcal{H}$: Human difficulty; $\mathcal{M}$: Model difficulty.
    Saturated represents the total ratio of items that 90\% of the model can solve correctly.
    Savant represents the ratio of items that are hard for humans (difficulty top 33\%) but trivial for most models (correct for 90\% of the models).
    Brittleness represents the ratio of items that are easy for humans, where most of the models fail.
    \textbf{The saturated rate and savant rate present the diagnosis of Human-LLM Difficulty Misalignment.}
    }
    \vspace{-2.2mm}
    \label{tab:table_irt}
    \end{table}

\begin{table*}[t]
    \centering
    \resizebox{\textwidth}{!}{
    \begin{tabular}{lcccccccccccccccc|c}
    \toprule
    & \multicolumn{4}{c}{\textbf{USMLE}} & \multicolumn{4}{c}{\textbf{Cambridge}} & \multicolumn{4}{c}{\textbf{SAT Reading\&Writing}} & \multicolumn{4}{c}{\textbf{SAT Math}} & \\
    \cmidrule(lr){2-5} \cmidrule(lr){6-9} \cmidrule(lr){10-13} \cmidrule(lr){14-17}
    \textbf{Model} & \textbf{Baseline} & \textbf{Weak} & \textbf{Medium} & \textbf{Strong} & \textbf{Baseline} & \textbf{Weak} & \textbf{Medium} & \textbf{Strong} & \textbf{Baseline} & \textbf{Weak} & \textbf{Medium} & \textbf{Strong} & \textbf{Baseline} & \textbf{Weak} & \textbf{Medium} & \textbf{Strong} & \textbf{Average} \\
    \midrule
    GPT-3.5-Turbo & 0.868 & 0.862\rlap{$\downarrow$} & 0.882 & 0.877\rlap{$\uparrow$} & 0.632 & 0.628\rlap{$\downarrow$} & 0.612 & 0.613 & 0.660 & 0.667 & 0.685 & 0.687\rlap{$\uparrow$} & 0.727 & 0.750 & 0.749 & 0.762\rlap{$\uparrow$} & 0.729 \\
    GPT-4o & 0.952 & 0.960 & 0.957 & 0.955\rlap{$\uparrow$} & 0.900 & 0.893\rlap{$\downarrow$} & 0.898 & 0.897 & 0.871 & 0.870\rlap{$\downarrow$} & 0.874 & 0.874\rlap{$\uparrow$} & 0.908 & 0.907\rlap{$\downarrow$} & 0.915 & 0.913\rlap{$\uparrow$} & 0.909 \\
    GPT-4o-mini & 0.925 & 0.928 & 0.933 & 0.934\rlap{$\uparrow$} & 0.786 & 0.765\rlap{$\downarrow$} & 0.783 & 0.793\rlap{$\uparrow$} & 0.780 & 0.786 & 0.787 & 0.783\rlap{$\uparrow$} & 0.899 & 0.898\rlap{$\downarrow$} & 0.898 & 0.901\rlap{$\uparrow$} & 0.849 \\
    GPT-4.1 & 0.895 & 0.906 & 0.906 & 0.898\rlap{$\uparrow$} & 0.919 & 0.922 & 0.922 & 0.918 & 0.888 & 0.915 & 0.913 & 0.905\rlap{$\uparrow$} & 0.904 & 0.905 & 0.896 & 0.910\rlap{$\uparrow$} & 0.908 \\
    GPT-4.1-mini & \textbf{0.966} & 0.939\rlap{$\downarrow$} & 0.966 & 0.957 & 0.897 & 0.868\rlap{$\downarrow$} & 0.895 & 0.880 & 0.895 & 0.892\rlap{$\downarrow$} & 0.907 & 0.904\rlap{$\uparrow$} & 0.922 & 0.920\rlap{$\downarrow$} & 0.927 & 0.925\rlap{$\uparrow$} & 0.916 \\
    GPT-o4-mini & 0.780 & 0.799 & 0.810 & 0.772 & 0.909 & 0.895\rlap{$\downarrow$} & 0.898 & 0.904 & 0.908 & 0.898\rlap{$\downarrow$} & 0.909 & 0.904 & 0.906 & 0.909 & 0.909 & 0.902 & 0.876 \\
    GPT-5 & 0.946 & 0.939\rlap{$\downarrow$} & 0.942 & 0.946 & \textbf{0.958} & 0.957\rlap{$\downarrow$} & 0.956 & 0.961\rlap{$\uparrow$} & \textbf{0.976} & 0.978 & 0.976 & 0.977\rlap{$\uparrow$} & 0.924 & 0.914\rlap{$\downarrow$} & 0.926 & 0.923 & \textbf{0.950} \\
    \midrule
    Llama2-7B & 0.712 & 0.718 & 0.687 & 0.700 & 0.397 & 0.397 & 0.405 & 0.409\rlap{$\uparrow$} & 0.396 & 0.380\rlap{$\downarrow$} & 0.401 & 0.392 & 0.230 & 0.224\rlap{$\downarrow$} & 0.221 & 0.222 & 0.431 \\
    Llama2-13B & 0.775 & 0.720\rlap{$\downarrow$} & 0.763 & 0.768 & 0.459 & 0.428\rlap{$\downarrow$} & 0.455 & 0.436 & 0.469 & 0.477 & 0.466 & 0.461 & 0.267 & 0.264\rlap{$\downarrow$} & 0.262 & 0.268\rlap{$\uparrow$} & 0.484 \\
    Phi3 & 0.844 & 0.843\rlap{$\downarrow$} & 0.843 & 0.844 & 0.493 & 0.489\rlap{$\downarrow$} & 0.489 & 0.504\rlap{$\uparrow$} & 0.632 & 0.662 & 0.638 & 0.649\rlap{$\uparrow$} & 0.774 & 0.777 & 0.788 & 0.770 & 0.690 \\
    Phi3.5 & 0.867 & 0.853\rlap{$\downarrow$} & 0.870 & 0.865 & 0.559 & 0.550\rlap{$\downarrow$} & 0.540 & 0.551 & 0.647 & 0.646\rlap{$\downarrow$} & 0.645 & 0.642 & 0.775 & 0.787 & 0.784 & 0.782\rlap{$\uparrow$} & 0.710 \\
    Llama3.1-8B & 0.891 & 0.897 & 0.892 & 0.880 & 0.646 & 0.633\rlap{$\downarrow$} & 0.649 & 0.639 & 0.685 & 0.683\rlap{$\downarrow$} & 0.674 & 0.674 & 0.766 & 0.751\rlap{$\downarrow$} & 0.778 & 0.771\rlap{$\uparrow$} & 0.744 \\
    Qwen2.5-7B & 0.850 & 0.856 & 0.847 & 0.852\rlap{$\uparrow$} & 0.675 & 0.686 & 0.663 & 0.690\rlap{$\uparrow$} & 0.730 & 0.719\rlap{$\downarrow$} & 0.727 & 0.736\rlap{$\uparrow$} & 0.895 & 0.900 & 0.897 & 0.894 & 0.789 \\
    Qwen2.5-32B & 0.922 & 0.931 & 0.921 & 0.933\rlap{$\uparrow$} & 0.783 & 0.812 & 0.812 & 0.803\rlap{$\uparrow$} & 0.812 & 0.813 & 0.819 & 0.816\rlap{$\uparrow$} & 0.919 & 0.917\rlap{$\downarrow$} & 0.920 & 0.921\rlap{$\uparrow$} & 0.866 \\
    Phi4 & 0.921 & 0.918\rlap{$\downarrow$} & 0.928 & 0.924\rlap{$\uparrow$} & 0.798 & 0.791\rlap{$\downarrow$} & 0.778 & 0.784 & 0.802 & 0.802 & 0.801 & 0.798 & 0.919 & 0.913\rlap{$\downarrow$} & 0.908 & 0.913 & 0.856 \\
    Qwen3-8B & 0.903 & 0.907 & 0.909 & 0.901 & 0.767 & 0.776 & 0.765 & 0.768\rlap{$\uparrow$} & 0.820 & 0.816\rlap{$\downarrow$} & 0.811 & 0.802 & 0.919 & 0.926 & 0.916 & 0.925\rlap{$\uparrow$} & 0.852 \\
    Qwen3-32B & 0.946 & 0.957 & 0.948 & 0.948\rlap{$\uparrow$} & 0.845 & 0.863 & 0.849 & 0.856\rlap{$\uparrow$} & 0.881 & 0.879\rlap{$\downarrow$} & 0.868 & 0.868 & 0.927 & 0.932 & 0.923 & 0.926 & 0.901 \\
    \midrule
    DeepSeek-R1 & 0.961 & 0.966 & 0.955 & \textbf{0.967}\rlap{$\uparrow$} & \underline{0.933} & 0.919\rlap{$\downarrow$} & 0.931 & 0.922 & \underline{0.964} & 0.959\rlap{$\downarrow$} & 0.968 & 0.954 & 0.914 & 0.919 & 0.915 & 0.913 & \underline{0.941} \\
    QWQ-32B & 0.954 & 0.948\rlap{$\downarrow$} & 0.960 & 0.954 & 0.898 & 0.890\rlap{$\downarrow$} & 0.899 & 0.899\rlap{$\uparrow$} & 0.921 & 0.922 & 0.915 & 0.920 & \textbf{0.953} & 0.943\rlap{$\downarrow$} & 0.943 & 0.941 & 0.929 \\
    R1-Qwen32B & 0.939 & 0.936\rlap{$\downarrow$} & 0.940 & 0.940\rlap{$\uparrow$} & 0.856 & 0.827\rlap{$\downarrow$} & 0.868 & 0.849 & 0.886 & 0.877\rlap{$\downarrow$} & 0.881 & 0.884 & 0.944 & 0.947 & 0.942 & 0.941 & 0.904 \\
    Qwen3-32B (R) & 0.954 & 0.963 & 0.969 & 0.966\rlap{$\uparrow$} & 0.907 & 0.904\rlap{$\downarrow$} & 0.898 & 0.897 & 0.932 & 0.917\rlap{$\downarrow$} & 0.926 & 0.925 & \underline{0.952} & 0.943\rlap{$\downarrow$} & 0.938 & 0.938 & 0.933 \\
    \bottomrule
    \end{tabular}
    }
    \vspace{-2.2mm}
    \caption{Problem-solving accuracy across different personas. An arrow $\uparrow$ or $\downarrow$ is added if the accuracy is improved or degraded compared to the baseline with the corresponding persona. \textbf{The magnitude of change is marginal, indicating that models struggle to significantly suppress or enhance their intrinsic capabilities on command.}}
    \label{tab:table_direct_results}
    \vspace{-3.2mm}
\end{table*}

\begin{findingbox}[title={Key Finding 2: Limits of Simulation.}]
    \textbf{Neither extrinsic ensembling nor intrinsic role-playing} serves as a reliable solution for alignment. We find that \textbf{ensemble performance is strictly bounded by individual model capabilities}: rather than contributing diverse insights, weaker models introduce noise that degrades the performance of high-capability baselines. Similarly, \textbf{proficiency simulation shows highly inconsistent results}, as models struggle to authentically mimic different proficiency levels or suppress their intrinsic knowledge: it fails as a faithful cognitive model, but proficiency ensembling can act as a variance-reduction heuristic.
\end{findingbox}

We first perform a greedy ensemble analysis, iteratively aggregating the predicted difficulty scores of the top-$K$ performing models. As shown in Figure \ref{fig:ensemble_topk}, the performance trend implies an \textbf{upper bound} for alignment capability that is strictly governed by the density of high-quality models within the pool. In domains with a dense cluster of capable models (SAT Math), the ensemble acts as a \textbf{denoising mechanism}, steadily improving correlation from $0.56$ to a peak of $0.66$. However, this gain is fragile: once the threshold ($K=14$) is crossed, adding weaker models causes immediate \textbf{signal dilution}. In sparse domains like \textbf{USMLE}, this dilution happens immediately, confirming that the long tail of weaker models introduces destructive noise rather than helpful diversity. Thus, extrinsic aggregation is not a scalable solution but a bounded analysis.

We further explore explicit cognitive simulation by prompting models to adopt specific student proficiencies. A granular analysis in Figure \ref{fig:hetmap_role_diff}, the changing ratio of utilizing different proficiencies, reveals that single-proficiency simulation is highly unstable: adopting a specific role often either improves or degrades alignment, suggesting models struggle to faithfully simulate a specific cognitive state in isolation.

However, Table \ref{tab:table_diff_avg_by_task} reveals another crucial nuance: while individual simulations are stochastic, the \textbf{ensemble average of proficiencies consistently improves alignment}. This benefit is particularly evident in frontier models; for instance, GPT-5 improves its average correlation to $0.47$ compared to its baseline of $0.34$. This suggests that while models cannot reliably act as a specific student, aggregating their intrinsic variations provides a more robust estimate than a single pass, effectively smoothing out the randomness of individual proficiency prompts. This improvement does not indicate successful student simulation, but rather reflects noise averaging over inconsistent internal states.

\section{Capability vs. Perception}

While the previous section examined the model as an external observer attempting to predict student struggles, we now turn our focus to the model as an internal actor. To understand the root causes of the observed misalignment, in this section, we disentangle the relationship between what models predict and what they experience.

\subsection{The Curse of Knowledge}

\begin{findingbox}[title={Key Finding 3: The Curse of Knowledge}]
    Our IRT-based analysis reveals that the difficulty derived from models' actual correctness correlates \textbf{even worse with human than what they explicitly perceive}. This stems from a distinct mechanistic divergence: \textbf{items that are conceptually difficult for humans are frequently trivial for models}, resulting in high saturation rates on hard items. Furthermore, this capability exhibits significant inertia: even when explicitly prompted to simulate a lower-proficiency student, \textbf{models fail to meaningfully suppress their problem-solving capability}.
\end{findingbox}

We shift our perspective from \textit{prediction} to \textit{experience} by treating our suite of 21 models as a cohort of students to estimate intrinsic machine difficulty (b parameter) via Item Response Theory (IRT).
In addition, to further quantify and analyze the divergence, we define additional metrics to be conditional on human difficulty tiers: the \textit{Savant Rate} measures the percentage of items within the top 33\% of human difficulty that are solved by over 90\% of models, while the \textit{Brittleness Rate} measures the percentage of items within the bottom 33\% of human difficulty where the model pass rate is below 50\%.

Table \ref{tab:table_irt} exposes a profound \textbf{Cognitive Divergence}, where the correlation of model IRT difficulty is even lower than the model perception. In domains like \textbf{USMLE}, the misalignment is extremely discouraging.
The Savant Rate of $70.4\%$ implies that for over two-thirds of the questions, humans find most difficult, can be solved by over 90\% of models easily. Coupled with a massive dataset-wide saturation of $75.6\%$, this creates a flat difficulty that explains the poor alignment.

In contrast, reasoning domains like \textbf{SAT Math} show partial alignment but remain limited by hyper-competence. While the correlation ($\rho=0.386$) is higher than in knowledge domains, the Savant Rate is still significant at $32.2\%$. It indicates that even in logical reasoning, one-third of the problems that challenge humans are trivial for machines. Furthermore, with $54.6\%$ of all items being saturated, models are getting too powerful capabilities to simulate student struggles.

Finally, we examine whether simulating proficiencies shifts intrinsic capability. As shown in Table \ref{tab:table_direct_results}, while applying a proficiency generally shifts accuracy in the expected direction, the magnitude is negligible (typically $<1\%$ change). This reveals the \textbf{Curse of Knowledge}. Highly capable models are too strong to fail, and their intrinsic objective to maximize correctness overrides proficiency instructions. This prevents them from authentically simulating the specific misconceptions of struggling students, as they cannot unsee the correct answers they already possess.

\begin{table}[t]
    \centering
    \setlength{\tabcolsep}{2.5pt}

    \resizebox{\linewidth}{!}{
        \begin{tabular}{l|cccc|c}
        \toprule
        \textbf{Model} & \textbf{USMLE} & \textbf{Cambridge} & \textbf{SAT-R} & \textbf{SAT-M} & \textbf{Average} \\
        \midrule
        GPT-3.5-Turbo   & 0.54 & 0.51 & 0.49 & \textbf{0.65} & 0.55 \\
        GPT-4o          & 0.62 & 0.59 & 0.54 & 0.63 & 0.60 \\
        GPT-4o-mini     & 0.55 & 0.57 & 0.51 & 0.63 & 0.56 \\
        GPT-4.1         & 0.55 & \underline{0.61} & 0.58 & 0.57 & 0.57 \\
        GPT-4.1-mini    & 0.55 & 0.56 & 0.59 & 0.62 & 0.58 \\
        GPT-o4-mini     & 0.52 & 0.59 & 0.55 & 0.53 & 0.55 \\
        GPT-5           & 0.60 & \textbf{0.73} & \textbf{0.72} & 0.60 & \textbf{0.67} \\
        \midrule
        Llama2-7B       & 0.49 & 0.50 & 0.46 & 0.54 & 0.50 \\
        Llama2-13B      & 0.48 & 0.49 & 0.50 & 0.54 & 0.50 \\
        Phi3            & 0.47 & 0.48 & 0.49 & 0.62 & 0.51 \\
        Phi3.5          & 0.50 & 0.51 & 0.48 & 0.58 & 0.52 \\
        Llama3.1-8B     & 0.53 & 0.51 & 0.49 & 0.63 & 0.54 \\
        Qwen2.5-7B      & 0.53 & 0.52 & 0.47 & 0.57 & 0.52 \\
        Qwen2.5-32B     & 0.55 & 0.54 & 0.49 & 0.62 & 0.55 \\
        Phi4            & 0.53 & 0.54 & 0.57 & \textbf{0.65} & 0.57 \\
        Qwen3-8B        & 0.55 & 0.52 & 0.49 & \underline{0.64} & 0.55 \\
        Qwen3-32B       & 0.62 & 0.52 & 0.52 & 0.61 & 0.57 \\
        \midrule
        DeepSeek-R1     & 0.64 & 0.53 & \underline{0.64} & 0.62 & \underline{0.61} \\
        QWQ-32B         & \underline{0.65} & 0.58 & 0.59 & \textbf{0.65} & 0.62 \\
        R1-Qwen32B      & 0.58 & 0.54 & 0.55 & \textbf{0.65} & 0.58 \\
        Qwen3-32B (R)   & \textbf{0.69} & 0.56 & 0.54 & 0.58 & 0.59 \\
        \midrule
        Average         & 0.55 & 0.55 & 0.54 & 0.60 & 0.56 \\
        \bottomrule
        \end{tabular}
    }
    \caption{
        AUROC (Area Under the ROC Curve) that measures the alignment between predicted difficulty and the model's own correctness. A value of 0.5 indicates random alignment. \textbf{Most models hover near 0.55, revealing a critical lack of self-awareness: they fail to predict their own potential errors.}
        }
    \vspace{-4.2mm}
    \label{tab:table_auroc}
\end{table}

\subsection{Metacognitive Blindness.}

\begin{findingbox}[title={Key Finding 4: Metacognitive Blindness.}]
    Our AUROC analysis reveals a critical lack of introspection: models are unable to accurately predict their own potential limitations. With AUROC scores hovering near random guessing (approximately 0.55) across most models, we find that \textbf{explicit difficulty estimates are effectively decoupled from the model's actual correctness}. This suggests a fundamental blind spot: because models cannot reliably identify which tasks exceed their own capabilities, they lack the necessary internal signal to ground their estimates of human difficulty.
\end{findingbox}

A fundamental question remains regarding whether the model's perception of difficulty is aligned with its own problem-solving capability. To quantify this \textit{Metacognition}, we formulate difficulty prediction as a binary classification task regarding the model's own correctness.

For a given item $x_i$, we assign a label $l_i = 1$ if the model answers incorrectly ($v_i = 0$) and $l_i = 0$ if it answers correctly ($v_i = 1$), using the predicted difficulty score $\hat{y}_i$ as the prediction probability. We calculate the Area Under the Receiver Operating Characteristic Curve (AUROC) to measure separability, where the value represents \textit{the probability that the model assigns a higher difficulty score to an item it answers incorrectly compared to one it answers correctly}. An AUROC of approximately $0.5$ indicates random alignment where the model lacks self-awareness, whereas values significantly greater than $0.5$ demonstrate positive metacognition where the model correctly identifies its failure.

Table \ref{tab:table_auroc} presents the results of this analysis and exposes a critical \textbf{Metacognitive Blind Spot} across the majority of evaluated LLMs. Despite achieving high problem-solving accuracy, most models exhibit AUROC scores ranging between $0.50$ and $0.60$. This weak internal alignment implies that they lack the introspection to identify when a task exceeds their capabilities. Furthermore, even frontier models fail to demonstrate robust metacognition. While GPT-5 and DeepSeek-R1 show slight deviations from random guessing by achieving localized highs of $0.73$ on Cambridge and $0.64$ on USMLE, respectively, their overall discrimination remains poor, with averages hovering near or below $0.67$. These values indicate that even the most advanced models still \textbf{struggle to reliably distinguish between questions they can answer and those they cannot, highlighting that accurate self-awareness remains a persistent deficiency in current LLMs.}

\section{Conclusion}

This study demonstrates that Large Language Models currently struggle to align with human perception of difficulty despite their advanced problem-solving capabilities. We find that increasing model scale does not guarantee better alignment but rather fosters a machine consensus that systematically diverges from student reality. Our investigation attributes this failure to a fundamental capability gap where models cannot effectively suppress their knowledge to simulate struggling students, coupled with a critical lack of metacognitive introspection regarding their own limitations. Ultimately, these results highlight that bridging the gap between solving a problem and estimating its difficulty requires more than just stronger models or proficiency prompting, calling for new approaches to ground machine cognition in human educational needs.

\clearpage
\section*{Limitations}

Our investigation into proficiency simulation relied on zero-shot prompting strategies. While this reflects the most common and accessible usage of LLMs, it assumes that models can internally calibrate to a proficiency level without examples. We did not explore few-shot prompting with real student error patterns or fine-tuning on student response logs (Student Trace Modeling). It is possible that few-shot scenarios can improve the correlation between the Human-AI alignment. However, in this paper, we focus on the intrinsic capability of LLMs that are not affected by further learning or training processes.

\bibliography{custom}

\clearpage
\appendix
\startcontents[appendix]
\printcontents[appendix]{ }{0}{\section*{Table of Contents for Appendix}}

\clearpage
\section{Related Work}
\label{sec:related_work}

\subsection{Item Difficulty Prediction}

Conventional item difficulty prediction in large-scale assessments typically depends on item response data collected through field testing, where newly created items are embedded in operational forms (but not scored) and then analyzed within classical test theory (CTT) or item response theory (IRT) frameworks~\citep{hsu2018automated, benedetto2023quantitative}. Within these frameworks, CTT operationalizes difficulty as the proportion correct (p-value), whereas IRT models the probability of a correct response as a function of latent ability and item parameters via statistical models such as logit/probit links~\citep{demars2010item, hsu2018automated}. Despite their accuracy, field-testing and calibration are frequently criticized as time-consuming and costly, because data collection can take several months and IRT calibration may require administering items to several thousand examinees~\citep{ hambleton1991fundamentals, hsu2018automated, alkhuzaey2024text}. Embedding non-scored pretest items into operational tests also lengthens administrations and raises concerns about test-taker engagement and item exposure, which can compromise test security in high-stakes contexts~\citep{loukina2016textual, hsu2018automated, benedetto2023quantitative}. Expert-based difficulty ratings have been proposed as an alternative, but they are seldom used at scale due to subjectivity and weak alignment with psychometric difficulty, with reported evidence suggesting weak alignment between expert judgments and IRT-based difficulty estimates~\citep{conejo2014empirical}. In response, text-based item difficulty modeling predicts difficulty directly from item text and related metadata using machine learning, thereby avoiding response-data collection and potentially reducing time, cost, and reliance on subjective ratings~\citep{sano2015automated, loukina2016textual,huang2017question, hsu2018automated}. Early text-based work was dominated by feature engineering grounded in linguistic, cognitive, or psychometric theory, exemplified by combining linguistic indicators and item-level metadata, but such approaches can require substantial manual extraction and may generalize poorly across domains~\citep{perkins1995predicting, loukina2016textual}.

The learned representations \citep{devlin2019bert} are typically combined with regressors/classifiers such as CNNs or LSTMs for prediction, with empirical results showing that a CNN yielded superior performance to both a regular CNN variant and a TF-IDF+SVM baseline on TOEFL reading comprehension items~\citep{he2021automatically}. More recently, transformer-based small language models have been fine-tuned end-to-end for item difficulty prediction, with BERT first explored in large-scale assessment settings in 2021 and subsequent work showing that fine-tuned BERT and DistilBERT can outperform traditional linguistic/readability features and TF-IDF or Word2Vec-based approaches, while DistilBERT can match BERT performance at lower cost~\citep{mccarthy2021jump, benedetto2023quantitative}. In parallel, large language models have been used to directly predict difficulty or to generate auxiliary inputs and features, such as rationales, predicted answers, reasoning steps, uncertainty proxies, or simulated test-taker behaviors, that are then fed into downstream models for difficulty estimation~\citep{rogoz2024unibucllm, li2025item, feng2025reasoning, zotos2024you, duenas2024upn}. Across this section, the choice of modeling paradigm reflects a recurring trade-off between predictive power and interpretability, and some studies report that hand-crafted linguistic and metadata features can contribute more than BERT embeddings in specific settings~\citep{tack2024itec}.

\subsection{Student Simulation}

Student simulation refers to the use of artificial agents to generate learner-like behavior or data (e.g., behavioral traits, performance patterns, and learning progression), and is conceptually distinguished from student modeling systems that primarily infer latent states for adaptation~\citep{chrysafiadi2013student}. Early work emphasized three enduring motivations for student simulation: enabling teachers to ``practice the art of tutoring,'' supporting learning-by-teaching with a simulated peer, and allowing formative evaluation of instructional materials without relying entirely on human learner data~\citep{vanlehn1994applications}. Before the LLM era, building such simulations often required extensive hand-crafting of dialogue moves and misconception models in rule-based tutoring systems~\citep{graesser2005autotutor}, or large-scale human role-play data collection for tutoring dialogues~\citep{stasaski2020more}. With the advent of large language models, student simulation has been reframed as more feasible and scalable, driven by capabilities such as reproducing population-level behavioral distributions~\citep{weber2024constrained}, expressing reasoning/misunderstanding in natural language~\citep{zhang2021ai}, and supporting more agent-like behavior via memory and planning~\citep{park2023generative}. Simulated students have been used across major contexts including data generation, teacher training, learning by teaching/collaboration, and content evaluation, illustrating their broad educational utility~\citep{kaser2024simulated}.

LLM-based student simulation methods vary with the simulation goal, including modeling student traits, performance patterns, or learning progression, and are commonly instantiated via prompting, fine-tuning on learner traces, and agentic designs that support memory and planning~\citep{corbett1994knowledge, rasch1993probabilistic, lord2012applications, embretson2013item}. Prompt-based role conditioning remains a common baseline, where the simulator is instructed to ``act as'' a student with specified demographic/background/affective attributes~\citep{markel2023gpteach, lee2023generative, lusimstudent2024}, though direct prompting can struggle to reliably instantiate the intended persona characteristics. To better capture authentic learning dynamics (e.g., iterative struggle and revision), some approaches fine-tune LLMs on real student submission trajectories and temporally ordered traces~\citep{miroyan2025parastudent, ross2025modeling}. In parallel, agentic designs extend simulators beyond static role-play by endowing them with memory/reflection/decision-making components, building on the broader notion of LLM-based generative agents~\citep{park2023generative}.
Cognitive grounding can be strengthened by integrating knowledge tracing or IRT as internal state mechanisms to control and drive simulated behavior over time (e.g., evolving mastery/ability and its interaction with task difficulty), rather than only relying on surface-level role-play~\citep{corbett1994knowledge,rasch1993probabilistic,embretson2013item}. Despite these advances, the prior studies explicitly caution that fluent language output does not guarantee behavioral fidelity and note that many simulated-learner studies lack empirical realism validation, motivating more standardized and context-aware evaluation practices~\citep{kaser2024simulated}.
Similarly, \citet{park-etal-2024-large} introduce LLaSA, a framework that selects clusters of heterogeneous LLMs to approximate students at different ability levels and then applies IRT to the resulting synthetic response matrix for question difficulty estimation. This work provides evidence that LLM response patterns can be useful for psychometric estimation, but its calibrated setting depends on student response data for ability matching, leaving open whether LLMs independently align with human difficulty without such calibration.
\citet{hayakawa2024can} examine a closely related setting in second-language reading comprehension, where LLMs are prompted or perturbed to mimic CEFR-level language learners. They compare LLM next-token option distributions with human learner response distributions on CMCQRD and find that weakening prompts can reduce model performance, but do not reliably reproduce learner-like error patterns. This finding cautions that surface-level degradation or distributional smoothing is insufficient evidence of faithful student simulation.
\citet{liu2025leveraging} propose leveraging LLMs as synthetic examinees to estimate item difficulty through IRT calibration, demonstrating that model-generated responses can approximate human-derived item parameters under certain conditions. In contrast, our work does not treat alignment as a purely psychometric estimation problem, but instead interrogates the cognitive validity of such alignment by disentangling difficulty perception, intrinsic capability, and metacognitive awareness.
\citet{srivatsa-etal-2025-llms} find that strong general-purpose models consistently outperform average students without guidance, while weaker or domain-mismatched models may align only incidentally. Moreover, grade-enforcement prompts change model behavior but do not yield reliable alignment across models, prompts, subjects, and grades. These results highlight the need to validate whether simulated students are behaviorally faithful rather than assuming that role prompts or model capability imply student-like cognition.
\citet{sauberli-etal-2025-llms} find that temperature scaling improves distributional similarity but does not yield reliable alignment with human item facility, IRT-based response expectations, or distractor choices, suggesting that LLMs should not be directly treated as zero-shot pilot participants for assessment development.

\subsection{LLM Self-Awareness}

Numerous benchmarks study model self-knowledge through abstention, but they typically instantiate abstention under a single failure mode, such as unanswerable questions \citep{yin2023selfaware,amayuelas2024kuq}, multiple-choice questions with no correct option \citep{madhusudhan2025llms}, or underspecified inputs \citep{slobodkin2023musique-nq,zhang2024clamber,li2025questbench}. Closely related, verbalized uncertainty elicits a model's explicit expression of doubt and uses it as a downstream signal of whether the model can answer appropriately \citep{lin2022teaching,tian2023just}. However, several studies report that verbalized uncertainty can be brittle and may generalize poorly as a practical uncertainty-quantification mechanism \citep{vashurin2024benchmarking,lin2022teaching,xiong2024can}. At the same time, other work shows that these signals can be improved: \citet{kapoor2024large} demonstrates that fine-tuning can strengthen verbalized uncertainty, while \citet{kadavath2022language} shows that suitable prompting can elicit explicit correctness probabilities that become increasingly calibrated as model scale increases. Beyond uncertainty elicitation, abstention behavior itself has been improved via fine-tuning \citep{chen2024teaching,brahman2024art} and via explanation generation to justify refusals \citep{deng2024dontjustsay}, and related benchmarks also evaluate policy compliance and safety-related refusal behavior \citep{brahman2024art,DBLP:conf/aies/MuellerGBPP24,mazeika2024harmbench}. With the recent surge of interest in large reasoning models, MiP-Overthinking further reports that longer “thinking” traces may not improve abstention on unsolvable questions and can even exacerbate overconfident answering \citep{fan2025missing}. Complementing these lines of work, we study self-knowledge in the context of educational difficulty estimation: rather than only asking whether a model can abstain, we test whether its explicit difficulty judgments provide an introspective signal about its own likelihood of error on the same items. Our results indicate a pronounced metacognitive gap, difficulty estimates are only weakly predictive of failure cases, highlighting that off-the-shelf LLMs may lack the self-awareness needed to ground difficulty prediction in genuine student struggles.

\clearpage
\section{Consensus of Machines}
\label{sec:consensus_of_machines}

\begin{figure*}[htb]
    \centering
    \includegraphics[width=0.98\textwidth]{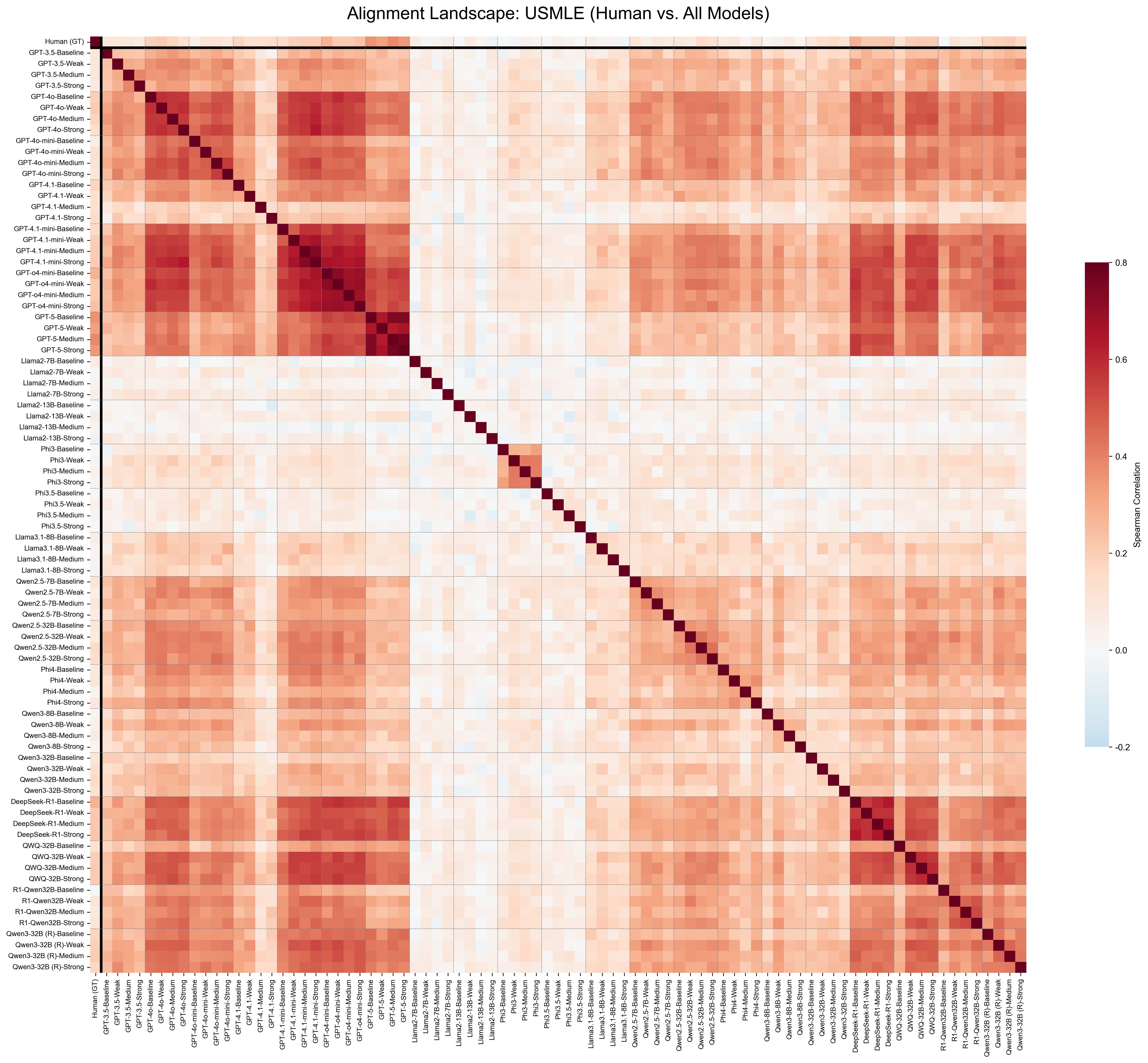}
    \vspace{-3.2mm}
    \caption{
        The consensus heatmap of the spearman correlation between the models on the USMLE dataset.
    }
    \vspace{-3.2mm}
    \label{fig:figure_appendix_consensus_USMLE}
    \end{figure*}

\begin{figure*}[htb]
    \centering
    \includegraphics[width=0.98\textwidth]{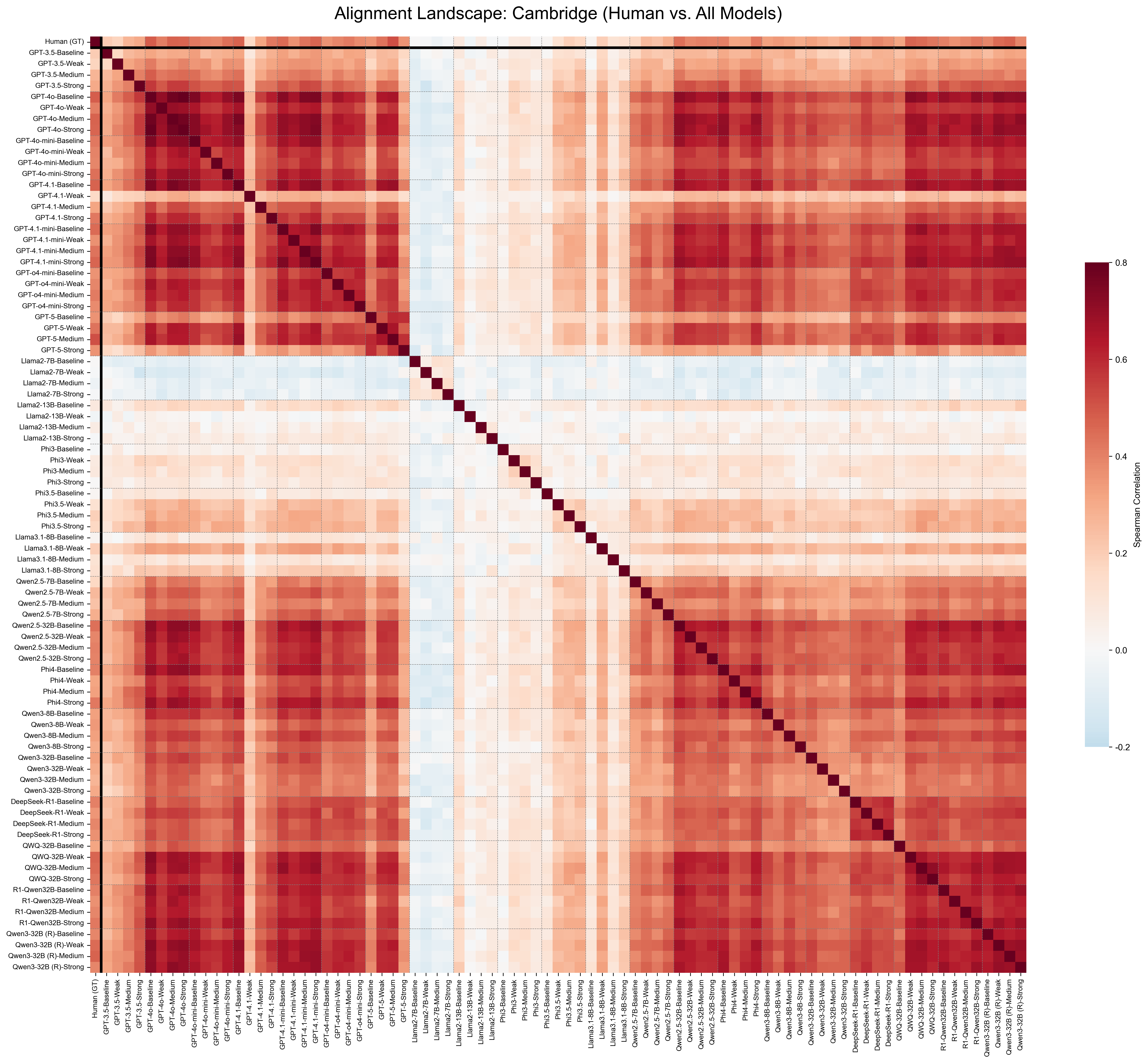}
    \vspace{-3.2mm}
    \caption{
        The consensus heatmap of the spearman correlation between the models on the CMCQRD dataset.
    }
    \vspace{-3.2mm}
    \label{fig:figure_appendix_consensus_Cambridge}
    \end{figure*}

\begin{figure*}[htb]
    \centering
    \includegraphics[width=0.98\textwidth]{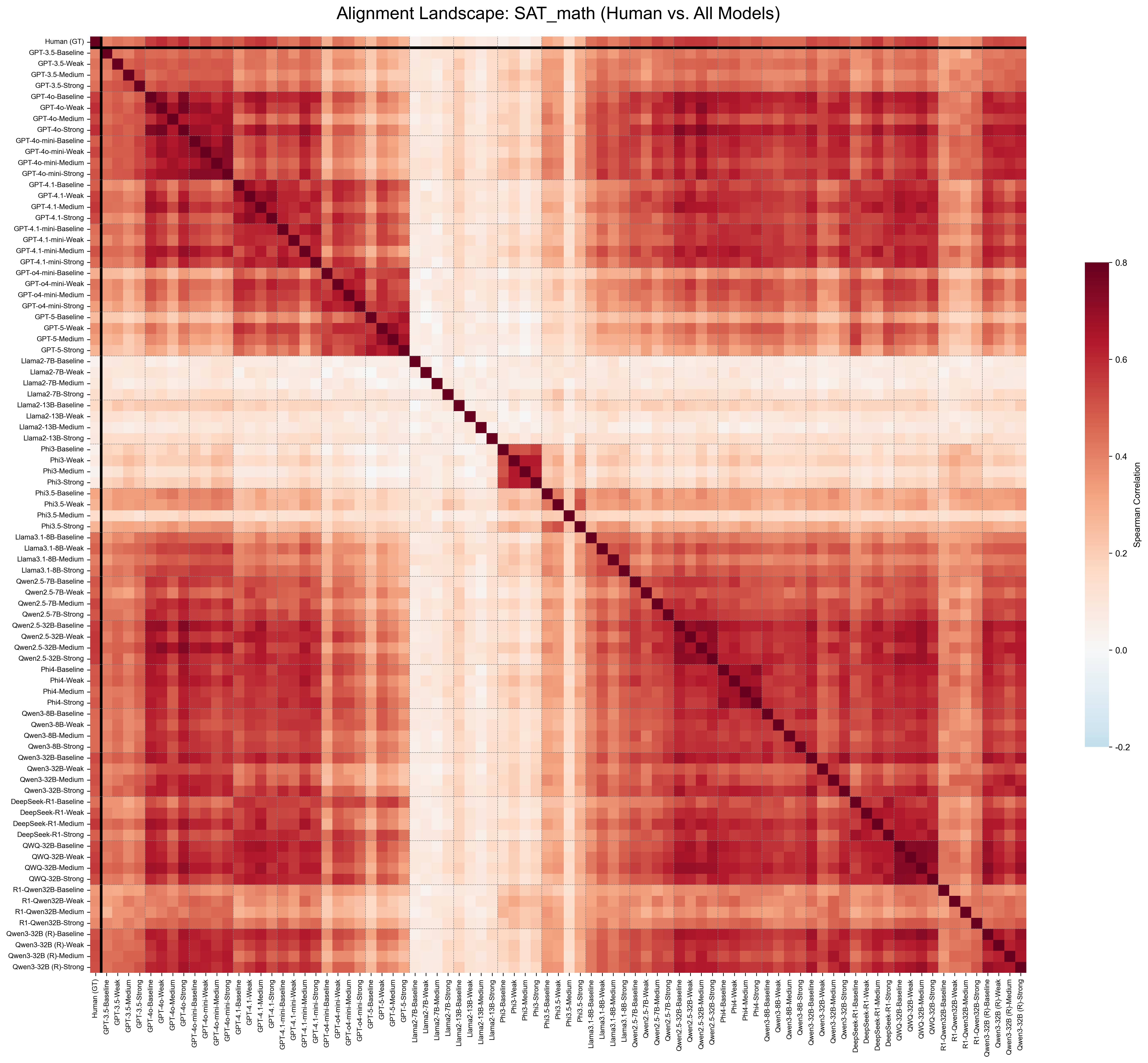}
    \vspace{-3.2mm}
    \caption{
        The consensus heatmap of the spearman correlation between the models on the SAT Math dataset.
    }
    \vspace{-3.2mm}
    \label{fig:figure_appendix_consensus_SAT_math}
    \end{figure*}

\begin{figure*}[htb]
    \centering
    \includegraphics[width=0.98\textwidth]{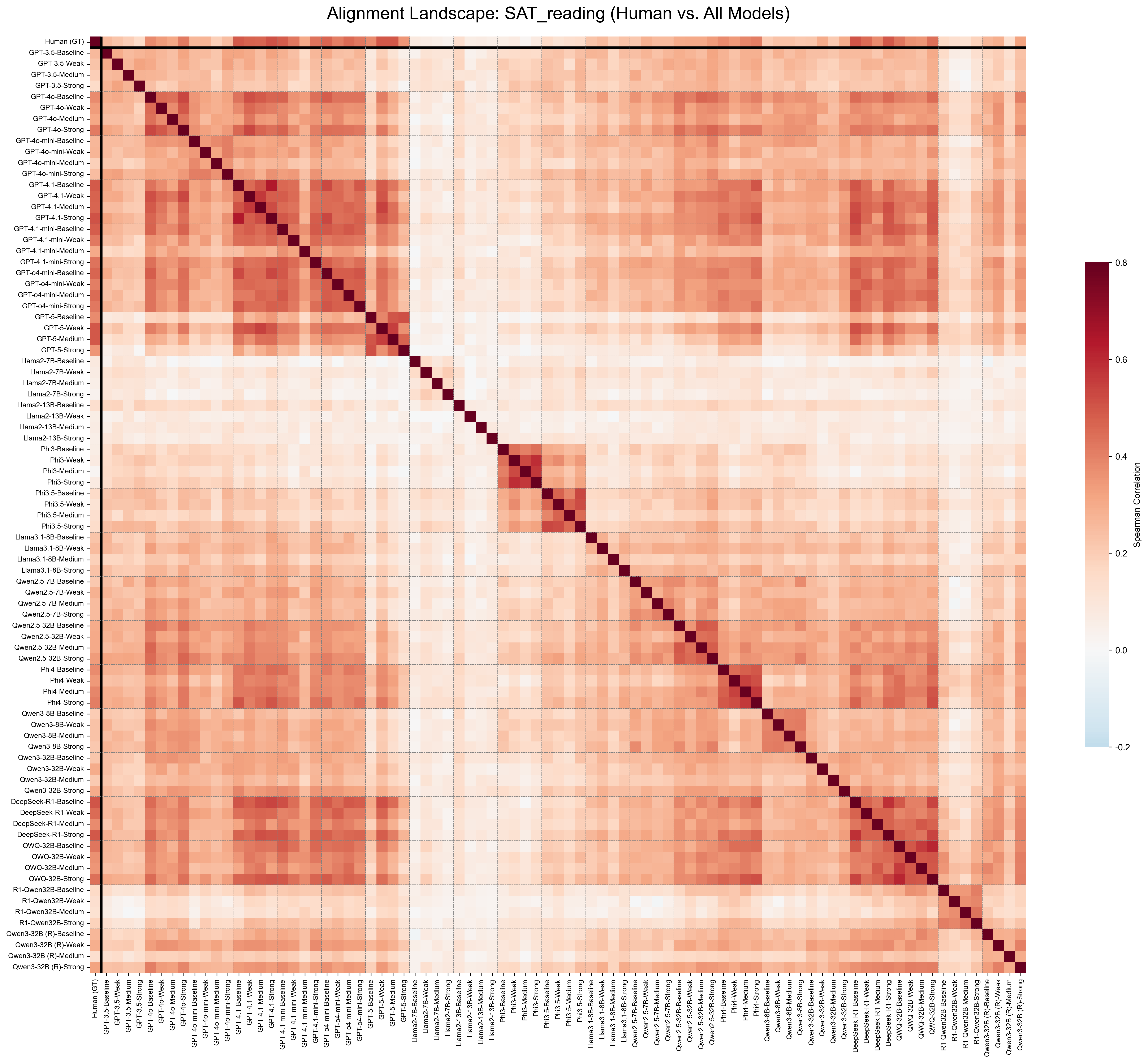}
    \vspace{-3.2mm}
    \caption{
        The consensus heatmap of the spearman correlation between the models on the SAT Reading dataset.
    }
    \vspace{-3.2mm}
    \label{fig:figure_appendix_consensus_SAT_reading}
    \end{figure*}

Figure \ref{fig:figure_appendix_consensus_USMLE}, \ref{fig:figure_appendix_consensus_Cambridge}, \ref{fig:figure_appendix_consensus_SAT_math}, \ref{fig:figure_appendix_consensus_SAT_reading} show the consensus heatmaps of the spearman correlation between the models on the USMLE, CMCQRD, SAT Math, and SAT Reading datasets. The heatmaps show that the correlation between the models is relatively higher than the correlation between the models and the human.


\clearpage
\section{Case Study}
\label{sec:case_study}

\begin{figure}[htb]
    \centering
    \includegraphics[width=0.46\textwidth]{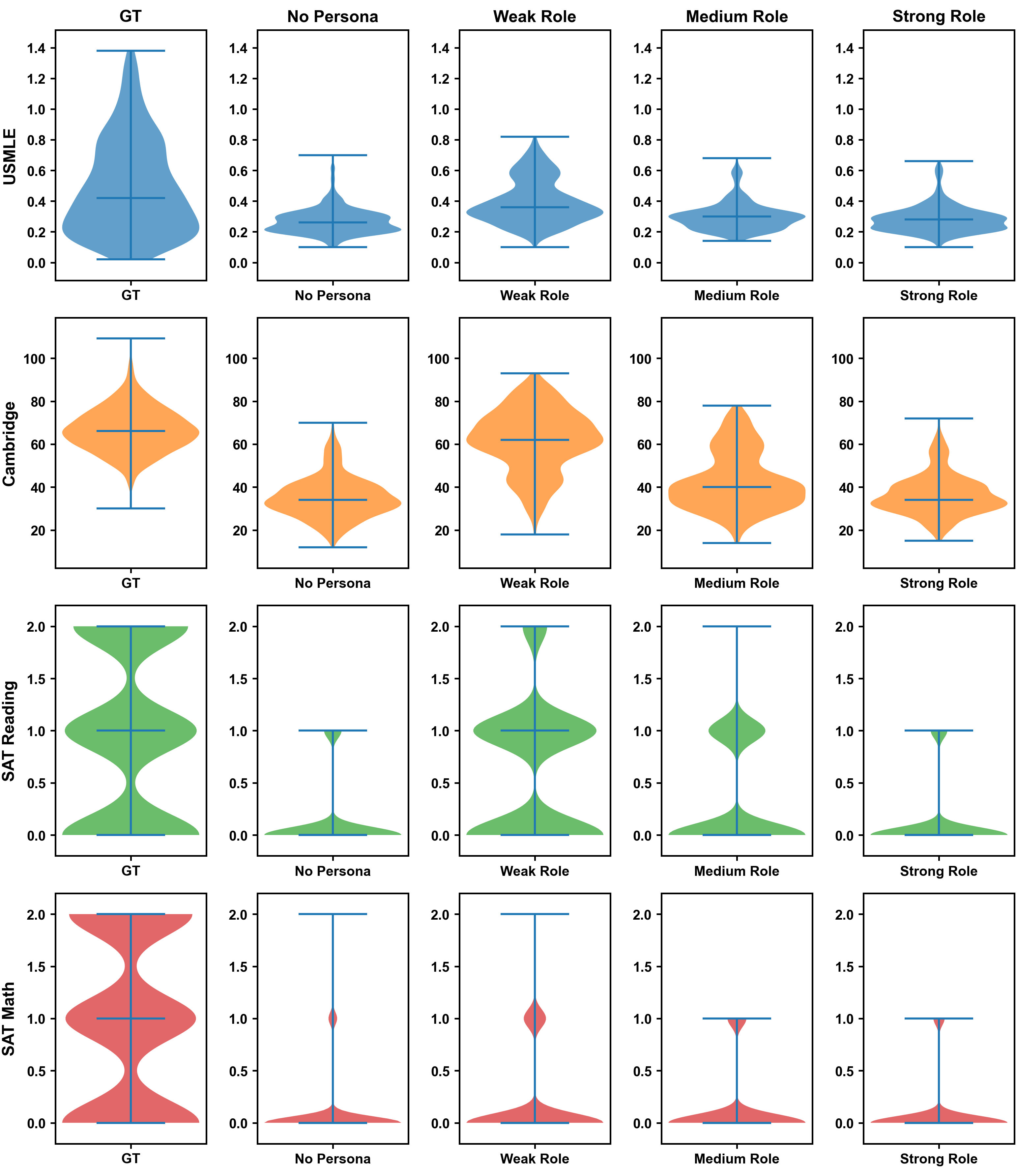}
    \caption{
    The violin plot of the difficulty prediction results of GPT-5 with different personas. \textbf{The Low-Proficiency simulation effectively expands the distribution, more closely resembling the Ground Truth. However, a systematic underestimation persists, indicating that the model remains optimistic about student performance compared to reality.}
    }
    \label{fig:diff_by_task_violin_gpt5}
    \end{figure}

Figure \ref{fig:diff_by_task_violin_gpt5} shows the violin plot of the difficulty prediction results of GPT-5 with different personas. The violin plot shows that the Low-Proficiency persona effectively expands the distribution, more closely resembling the Ground Truth. However, a systematic underestimation persists, indicating that the model remains optimistic about student performance compared to reality.

\clearpage
\section{Experimental Prompts}
\label{sec:experimental_prompts}

Figure \ref{fig:weak_student_diff_prompts}, \ref{fig:medium_student_diff_prompts}, \ref{fig:high_student_diff_prompts} show the prompt templates for difficulty prediction, including Low-Proficiency Student, Medium-Proficiency Student, and High-Proficiency Student across four tasks.
Figure \ref{fig:low_student_direct_prompts}, \ref{fig:medium_student_direct_prompts}, \ref{fig:high_student_direct_prompts} show the prompt templates for question answering, including Low-Proficiency Student, Medium-Proficiency Student, and High-Proficiency Student across four tasks.

\begin{figure*}[h]
    \centering
    \begin{tcolorbox}[
        colback=gray!5,
        colframe=black!45,
        fonttitle=\bfseries,
        title={Prompt for Difficulty Prediction: Low-Proficiency Student}
    ]
    \small
    \lstset{
        breaklines=true,
        basicstyle=\small\ttfamily,
        columns=fullflexible,
        keepspaces=true
    }
    \begin{lstlisting}
        ---------------------------------- For USMLE ---------------------------------

        (System Prompt)
        Suppose you are a student taking the USMLE exam. You are a weak student with low-level medical proficiency.

        (User Prompt)
        Analyze the difficulty values of the question. The difficulty values range from 0 to 1.0, where 0 is the easiest and 1.0 is the hardest. Analyze the difficulty step by step, and provide the final value in \boxed{...}:
        [Item Context]

        -------------------------------- For Cambridge -------------------------------

        (System Prompt)
        Suppose you are a student taking the Cambridge English Test. You are a weak student with low-level English proficiency.

        (User Prompt)
        Analyze the difficulty values of the question. The difficulty values range from 1 to 100, where 1 is the easiest and 100 is the hardest. Analyze the difficulty step by step, and provide the final value in \boxed{...}:
        [Item Context]

        ------------------------------ For SAT Reading -------------------------------

        (System Prompt)
        Suppose you are a student taking the SAT Reading exam. You are a weak student with low-level English proficiency.

        (User Prompt)
        Analyze the difficulty levels of the question. The difficulty levels contain 3 categories: easy, medium, and hard. Analyze the difficulty step by step, and provide the final category in \boxed{...}:
        [Item Context]

        -------------------------------- For SAT Math --------------------------------

        (System Prompt)
        Suppose you are a student taking the SAT math exam. You are a weak student with low-level math proficiency.

        (User Prompt)
        Analyze the difficulty levels of the question. The difficulty levels contain 3 categories: easy, medium, and hard. Analyze the difficulty step by step, and provide the final category in \boxed{...}:
        [Item Context]

    \end{lstlisting}
    \end{tcolorbox}
    \caption{Prompt templates for difficulty prediction (Low-Proficiency Student) across four tasks.}
    \label{fig:weak_student_diff_prompts}
    \end{figure*}

\begin{figure*}[h]
    \centering
    \begin{tcolorbox}[
        colback=gray!5,
        colframe=black!45,
        fonttitle=\bfseries,
        title={Prompt for Difficulty Prediction: Medium-Proficiency Student}
    ]
    \small
    \lstset{
        breaklines=true,
        basicstyle=\small\ttfamily,
        columns=fullflexible,
        keepspaces=true
    }
    \begin{lstlisting}
        ---------------------------------- For USMLE ---------------------------------

        (System Prompt)
        Suppose you are a student taking the USMLE exam. You are an average student with medium-level medical proficiency.

        (User Prompt)
        Analyze the difficulty values of the question. The difficulty values range from 0 to 1.0, where 0 is the easiest and 1.0 is the hardest. Analyze the difficulty step by step, and provide the final value in \boxed{...}:
        [Item Context]

        -------------------------------- For Cambridge -------------------------------

        (System Prompt)
        Suppose you are a student taking the Cambridge English Test. You are an average student with medium-level English proficiency.

        (User Prompt)
        Analyze the difficulty values of the question. The difficulty values range from 1 to 100, where 1 is the easiest and 100 is the hardest. Analyze the difficulty step by step, and provide the final value in \boxed{...}:
        [Item Context]

        ------------------------------ For SAT Reading -------------------------------

        (System Prompt)
        Suppose you are a student taking the SAT Reading exam. You are an average student with medium-level English proficiency.

        (User Prompt)
        Analyze the difficulty levels of the question. The difficulty levels contain 3 categories: easy, medium, and hard. Analyze the difficulty step by step, and provide the final category in \boxed{...}:
        [Item Context]

        -------------------------------- For SAT Math --------------------------------

        (System Prompt)
        Suppose you are a student taking the SAT math exam. You are an average student with medium-level math proficiency.

        (User Prompt)
        Analyze the difficulty levels of the question. The difficulty levels contain 3 categories: easy, medium, and hard. Analyze the difficulty step by step, and provide the final category in \boxed{...}:
        [Item Context]

    \end{lstlisting}
    \end{tcolorbox}
    \caption{Prompt templates for difficulty prediction (Medium-Proficiency Student) across four tasks.}
    \label{fig:medium_student_diff_prompts}
    \end{figure*}

    \begin{figure*}[h]
        \centering
        \begin{tcolorbox}[
            colback=gray!5,
            colframe=black!45,
            fonttitle=\bfseries,
            title={Prompt for Difficulty Prediction: High-Proficiency Student}
        ]
        \small
        \lstset{
            breaklines=true,
            basicstyle=\small\ttfamily,
            columns=fullflexible,
            keepspaces=true
        }
        \begin{lstlisting}
            ---------------------------------- For USMLE ---------------------------------

            (System Prompt)
            Suppose you are a student taking the USMLE exam. You are a good student with high-level medical proficiency.

            (User Prompt)
            Analyze the difficulty values of the question. The difficulty values range from 0 to 1.0, where 0 is the easiest and 1.0 is the hardest. Analyze the difficulty step by step, and provide the final value in \boxed{...}:
            [Item Context]

            -------------------------------- For Cambridge -------------------------------

            (System Prompt)
            Suppose you are a student taking the Cambridge English Test. You are a good student with high-level English proficiency.

            (User Prompt)
            Analyze the difficulty values of the question. The difficulty values range from 1 to 100, where 1 is the easiest and 100 is the hardest. Analyze the difficulty step by step, and provide the final value in \boxed{...}:
            [Item Context]

            ------------------------------ For SAT Reading -------------------------------

            (System Prompt)
            Suppose you are a student taking the SAT Reading exam. You are a good student with high-level English proficiency.

            (User Prompt)
            Analyze the difficulty levels of the question. The difficulty levels contain 3 categories: easy, medium, and hard. Analyze the difficulty step by step, and provide the final category in \boxed{...}:
            [Item Context]

            -------------------------------- For SAT Math --------------------------------

            (System Prompt)
            Suppose you are a student taking the SAT math exam. You are a good student with high-level math proficiency.

            (User Prompt)
            Analyze the difficulty levels of the question. The difficulty levels contain 3 categories: easy, medium, and hard. Analyze the difficulty step by step, and provide the final category in \boxed{...}:
            [Item Context]

        \end{lstlisting}
        \end{tcolorbox}
        \caption{Prompt templates for difficulty prediction (High-Proficiency Student) across four tasks.}
        \label{fig:high_student_diff_prompts}
        \end{figure*}

\begin{figure*}[h]
    \centering
    \begin{tcolorbox}[
        colback=gray!5,
        colframe=black!45,
        fonttitle=\bfseries,
        title={Prompt for Question Answering: Low-Proficiency Student}
    ]
    \small
    \lstset{
        breaklines=true,
        basicstyle=\small\ttfamily,
        columns=fullflexible,
        keepspaces=true
    }
    \begin{lstlisting}
        ---------------------------------- For USMLE ---------------------------------

        (System Prompt)
        Suppose you are a student taking the USMLE exam. You are a weak student with low-level medical proficiency.

        (User Prompt)
        Answer the question below step by step, and provide the final answer in \boxed{...}:
        [Item Context]

        -------------------------------- For Cambridge -------------------------------

        (System Prompt)
        Suppose you are a student taking the Cambridge English Test. You are a weak student with low-level English proficiency.

        (User Prompt)
        Answer the question below step by step, and provide the final answer in \boxed{...}:
        [Item Context]

        ------------------------------ For SAT Reading -------------------------------

        (System Prompt)
        Suppose you are a student taking the SAT Reading exam. You are a weak student with low-level English proficiency.

        (User Prompt)
        Answer the question below step by step, and provide the final answer in \boxed{...}:
        [Item Context]

        -------------------------------- For SAT Math --------------------------------

        (System Prompt)
        Suppose you are a student taking the SAT math exam. You are a weak student with low-level math proficiency.

        (User Prompt)
        Answer the question below step by step, and provide the final answer in \boxed{...}:
        [Item Context]

    \end{lstlisting}
    \end{tcolorbox}
    \caption{Prompt templates for question answering (Low-Proficiency Student) across four tasks.}
    \label{fig:low_student_direct_prompts}
    \end{figure*}

\begin{figure*}[h]
\centering
\begin{tcolorbox}[
    colback=gray!5,
    colframe=black!45,
    fonttitle=\bfseries,
    title={Prompt for Question Answering: Medium-Proficiency Student}
]
\small
\lstset{
    breaklines=true,
    basicstyle=\small\ttfamily,
    columns=fullflexible,
    keepspaces=true
}
\begin{lstlisting}
    ---------------------------------- For USMLE ---------------------------------

    (System Prompt)
    Suppose you are a student taking the USMLE exam. You are an average student with medium-level medical proficiency.

    (User Prompt)
    Answer the question below step by step, and provide the final answer in \boxed{...}:
    [Item Context]

    -------------------------------- For Cambridge -------------------------------

    (System Prompt)
    Suppose you are a student taking the Cambridge English Test. You are an average student with medium-level English proficiency.

    (User Prompt)
    Answer the question below step by step, and provide the final answer in \boxed{...}:
    [Item Context]

    -------------------------------- For SAT Reading --------------------------------

    (System Prompt)
    Suppose you are a student taking the SAT Reading exam. You are an average student with medium-level English proficiency.

    (User Prompt)
    Answer the question below step by step, and provide the final answer in \boxed{...}:
    [Item Context]

    -------------------------------- For SAT Math --------------------------------

    (System Prompt)
    Suppose you are a student taking the SAT math exam. You are an average student with medium-level math proficiency.

    (User Prompt)
    Answer the question below step by step, and provide the final answer in \boxed{...}:
    [Item Context]

\end{lstlisting}
\end{tcolorbox}
\caption{Prompt templates for question answering (Medium-Proficiency Student) across four tasks.}
\label{fig:medium_student_direct_prompts}
\end{figure*}

\begin{figure*}[h]
    \centering
    \begin{tcolorbox}[
        colback=gray!5,
        colframe=black!45,
        fonttitle=\bfseries,
        title={Prompt for Question Answering: High-Proficiency Student}
    ]
    \small
    \lstset{
        breaklines=true,
        basicstyle=\small\ttfamily,
        columns=fullflexible,
        keepspaces=true
    }
    \begin{lstlisting}
        ---------------------------------- For USMLE ---------------------------------

        (System Prompt)
        Suppose you are a student taking the USMLE exam. You are a good student with high-level medical proficiency.

        (User Prompt)
        Answer the question below step by step, and provide the final answer in \boxed{...}:
        [Item Context]

        -------------------------------- For Cambridge -------------------------------

        (System Prompt)
        Suppose you are a student taking the Cambridge English Test. You are a good student with high-level English proficiency.

        (User Prompt)
        Answer the question below step by step, and provide the final answer in \boxed{...}:
        [Item Context]

        ------------------------------ For SAT Reading -------------------------------

        (System Prompt)
        Suppose you are a student taking the SAT Reading exam. You are a good student with high-level English proficiency.

        (User Prompt)
        Answer the question below step by step, and provide the final answer in \boxed{...}:
        [Item Context]

        -------------------------------- For SAT Math --------------------------------

        (System Prompt)
        Suppose you are a student taking the SAT math exam. You are a good student with high-level math proficiency.

        (User Prompt)
        Answer the question below step by step, and provide the final answer in \boxed{...}:
        [Item Context]

    \end{lstlisting}
    \end{tcolorbox}
    \caption{Prompt templates for question answering (High-Proficiency Student) across four tasks.}
    \label{fig:high_student_direct_prompts}
    \end{figure*}

\end{document}